\documentclass[sigconf]{acmart}

\usepackage{booktabs}
\usepackage{subfig}
\usepackage{multirow}
\usepackage{amsthm}
\usepackage{hyperref}
\usepackage{algorithm}
\usepackage{algorithmic}
\usepackage{bm}
\usepackage{color}
\usepackage{colortbl}
\AtBeginDocument{%
  }

\setcopyright{acmcopyright}
\copyrightyear{2023}
\acmYear{2023}
\setcopyright{acmlicensed}\acmConference[MM '23]{Proceedings of the 31st ACM International Conference on Multimedia}{October 29-November 3, 2023}{Ottawa, ON, Canada}
\acmBooktitle{Proceedings of the 31st ACM International Conference on Multimedia (MM '23), October 29-November 3, 2023, Ottawa, ON, Canada}
\acmPrice{15.00}
\acmDOI{10.1145/3581783.3611928}
\acmISBN{979-8-4007-0108-5/23/10}

\begin{document}

\title{PAIF:  Perception-Aware   Infrared-Visible Image Fusion for Attack-Tolerant  Semantic Segmentation}


\author{Zhu Liu}
\affiliation{%
	\institution{Dalian University of Technology \country{}}
}
\email{liuzhu@mail.dlut.edu.cn}

\author{Jinyuan Liu}
\affiliation{%
	\institution{Dalian University of Technology \country{}}
}
\email{atlantis918@hotmail.com}

\author{Benzhuang Zhang}
\affiliation{%
	\institution{Dalian University of Technology \country{}}
}
\email{zbz@mail.dlut.edu.cn}

\author{Long Ma}
\affiliation{%
	\institution{Dalian University of Technology \country{}}
}
\email{malone94319@gmail.com}

\author{Xin Fan}
\affiliation{%
	\institution{Dalian University of Technology \country{}}
}
\email{xin.fan@dlut.edu.cn}

\author{Risheng Liu*}
\affiliation{%
	\institution{Dalian University of Technology \country{}}
		\institution{Peng Cheng Laboratory \country{}}
}
\email{rsliu@dlut.edu.cn}
\makeatletter
\def\authornotetext#1{
	\if@ACM@anonymous\else
	\g@addto@macro\@authornotes{
		\stepcounter{footnote}\footnotetext{#1}}
	\fi}
\makeatother
\authornotetext{Corresponding author: Risheng Liu.}
\renewcommand{\shortauthors}{Zhu Liu et al.}

\begin{abstract}
Infrared and visible image fusion is a powerful technique that combines complementary information from different modalities for downstream semantic perception tasks. Existing learning-based   methods show remarkable performance, but are suffering from the inherent vulnerability of adversarial attacks, causing a significant decrease in accuracy. In this work, a perception-aware fusion framework is proposed to promote segmentation robustness in adversarial scenes. 
We first conduct systematic analyses about the components of image fusion, investigating the correlation with segmentation robustness under adversarial perturbations. Based on these analyses, we propose a harmonized architecture search with a decomposition-based structure to balance standard accuracy and robustness. We also propose an adaptive learning strategy to improve the parameter robustness of image fusion, which can learn effective feature extraction under diverse adversarial perturbations. Thus, the goals of image fusion (\textit{i.e.,} extracting complementary features from source modalities and defending attack) can be realized from the perspectives of architectural and learning strategies.
Extensive experimental results demonstrate that our scheme substantially enhances the robustness, with gains of 15.3\% mIOU of segmentation in the adversarial scene, compared with  advanced competitors. The source codes are available at \url{https://github.com/LiuZhu-CV/PAIF}.
\end{abstract}


\maketitle

\section{Introduction}

Infrared-visible image fusion~\cite{U2Fusion2020,UMFusion,zhang2021sdnet,jiang2022towards} has attracted widespread attention in recent years, which provides complementary scene descriptions by integrating the merits of distinct visual modalities. In concrete,
thermal radiation-sensitive objects captured from infrared images can break down the limitations of harsh conditions (\textit{e.g.,} illumination and smoke), but these images are suffered from low resolution. Visible ones with abundant texture details are friable with the 
illumination changes. Thus, as a fundamental technique, image fusion not only provides visual-pleasant inception but also facilitates wide semantic perception tasks (\textit{e.g.,} segmentation~\cite{SeaFusion,liu2022learning} and object detection~\cite{TarDAL,ma2022pia,sun2022drone}).


With the flourished development of deep learning, amounts of  approaches~\cite{sun2019rtfnet,ha2017mfnet,TarDAL,SeaFusion,liu2023bilevel} are proposed for infrared-visible semantic segmentation, achieving remarkable performance.
However, there is a lack of sufficient researches to explore the robustness against adversarial attacks. The unnoticeable perturbations are ensconced in both modalities under diverse degraded conditions, that vulnerably compromise the stability and accuracy of network estimation~\cite{dong2019evading,su2018robustness}. We argue that existing multi-modality perception networks rely on the design of specialized feature aggregation  from dual backbones, \textit{e.g.,} attentions~\cite{lasnet}, edge-guided fusion~\cite{zhou2021gmnet} and concatenation~\cite{ha2017mfnet} for boosting the standard accuracy. Due to the lack of consideration for adversarial attacks,  these schemes cannot effectively remove the perturbations based on feature fusion and lead to the parameter oscillation at the phase of adversarial training~\cite{gu2022segpgd}.

\begin{figure}[t]
	\centering \begin{tabular}{c}
		\includegraphics[width=0.49\textwidth]{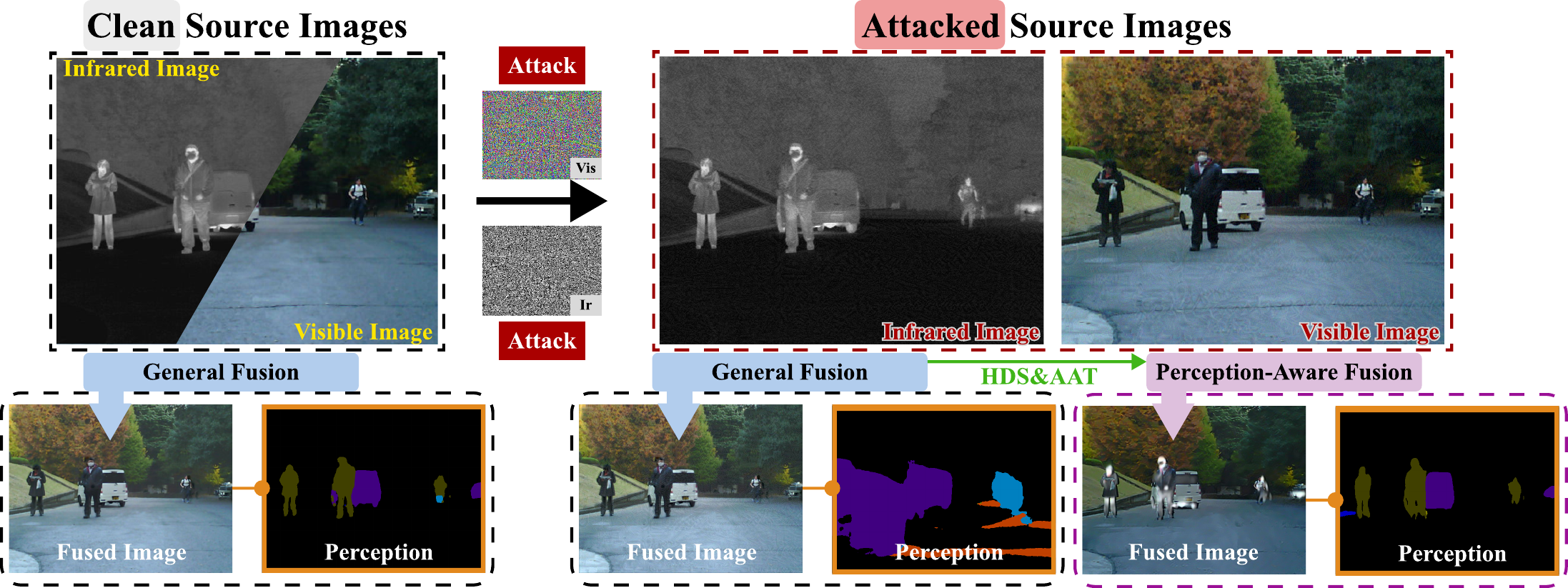}
	\end{tabular}
	\caption{Illustration of our core motivation that  to address the vulnerability of general image fusion for
		semantic segmentation under adversarial attacks, by introducing the proposed HDS and AAT, we construct the perception-aware  fusion for promoting attack tolerance of  segmentation.}
	\label{fig:first}
\end{figure}

Lately, the incorporated approaches of image fusion and semantic perception~\cite{TarDAL,SeaFusion,wu2022breaking,sun2022detfusion,wang2023interactively,liu2023task} are proposed to realize the well visual observation and scene understanding jointly. It furnishes a flexible way to obtain the perception-desired features for diverse multi-modality semantic understanding tasks, avoiding the complicated feature fusion mechanisms designs.
This empirical success motivates us
to investigate the robustness of image fusion for addressing the attacks of downstream perception, which is still blank, and limited by several challenging issues.

To be concrete, the major issue is  the methodology design of image fusion. Most approaches~\cite{U2Fusion2020,UMFusion,wu2022breaking,MFEIF} only focus on preserving the information of source images for visual inspection, such as texture details and thermal targets, while neglecting the intrinsic connection between fusion components and perception robustness. Secondly, as for fusion architectures, learning-based networks are mostly based on manual adjustments and abundant design experiences, ignoring the influences of adversarial attacks. Lately, as for the learning strategies,
current defence schemes concentrate on constructing specialized learning strategies and loss functions, like divide-and-conquer for segmentation~\cite{xu2021dynamic} and weighted class-wise learning for detection~\cite{chen2021class}. These approaches lack flexibility and cannot be easily generalized for  diverse vision perception tasks. Lately, various pre-processing techniques based on image enhancement (\textit{e.g.,} image super-resolution~\cite{yin2018deep,mustafa2019image}, dehazing~\cite{gao2021advhaze}, deraining~\cite{yu2022towards} and denoising~\cite{liao2018defense,xie2019feature}) are proposed to recover the images from adversarial corruptions and degradations. However, these methods cannot flexibly address various degrees of attacks.
In a word, our goal is to design a uniform scheme to investigate perception-aware image fusion against adversarial attacks of  semantic segmentation from both architectural design and learning strategy sides.










\begin{figure*}[thb]
	\centering \begin{tabular}{c@{\extracolsep{0.15em}}c@{\extracolsep{0.15em}}c@{\extracolsep{0.15em}}c}
		\includegraphics[width=0.24\textwidth]{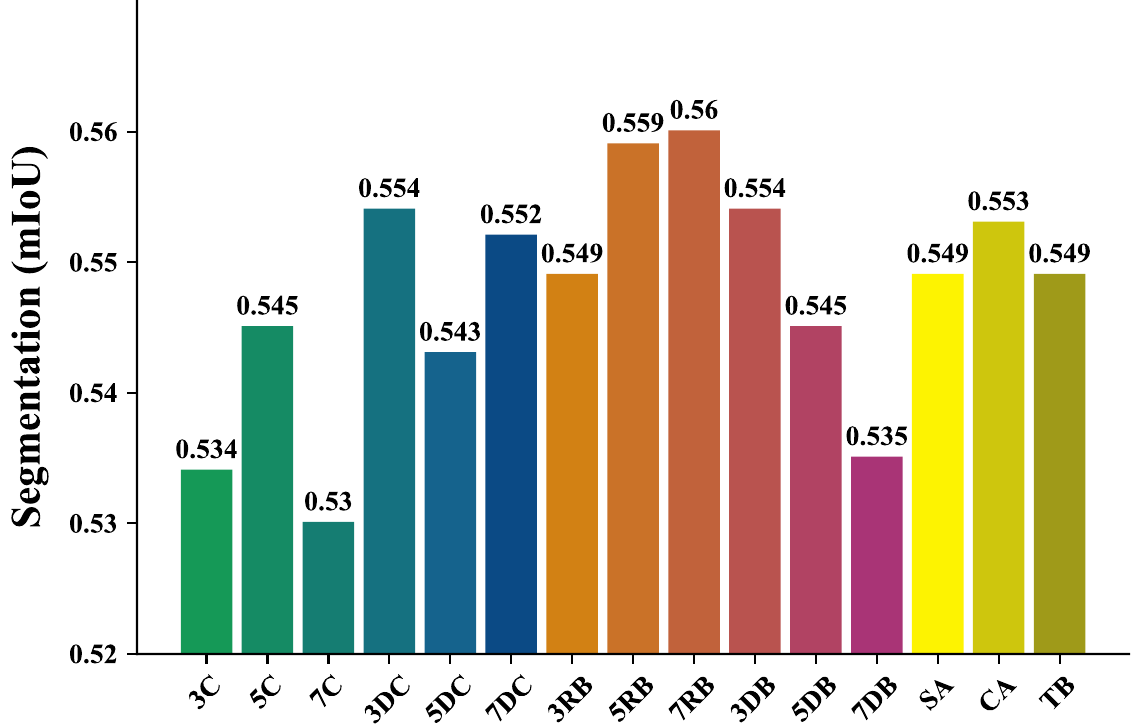}
		& \includegraphics[width=0.24\textwidth]{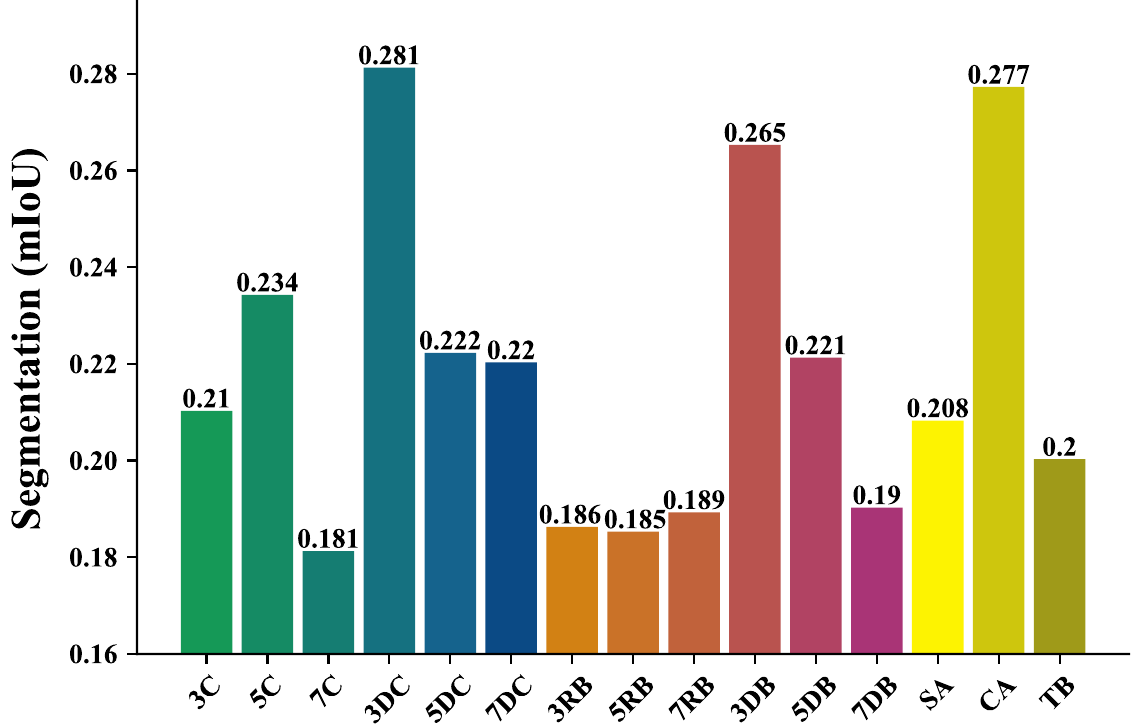}
		&		\includegraphics[width=0.24\textwidth]{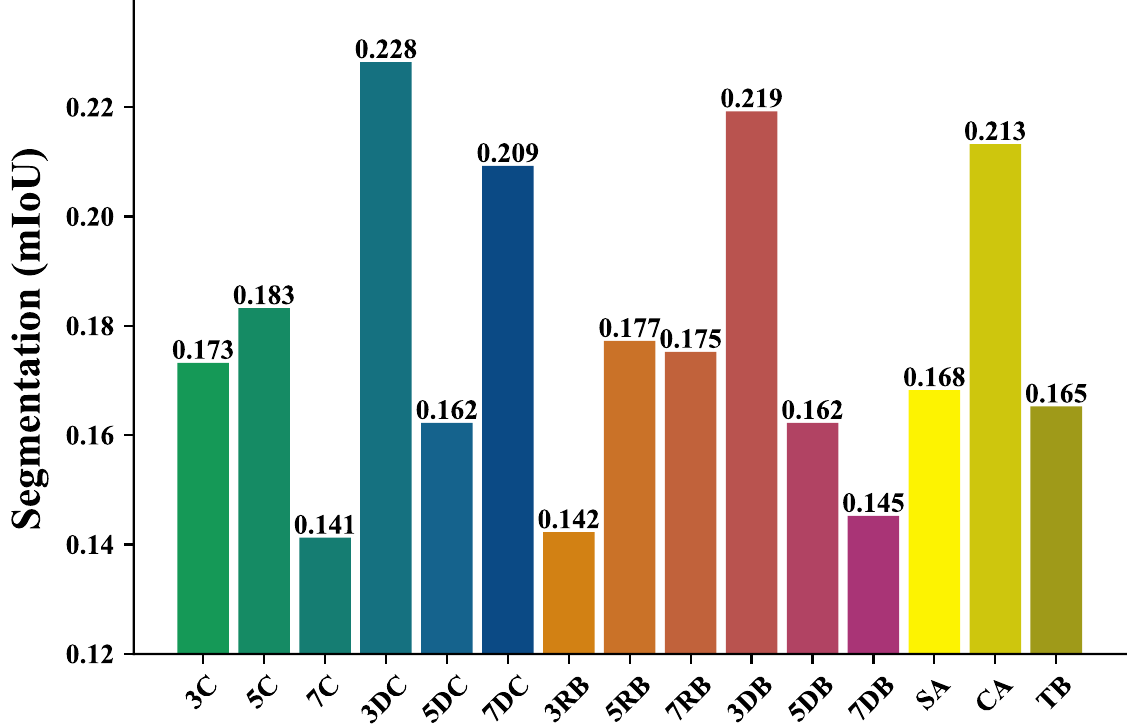}
		
		&		\includegraphics[width=0.24\textwidth]{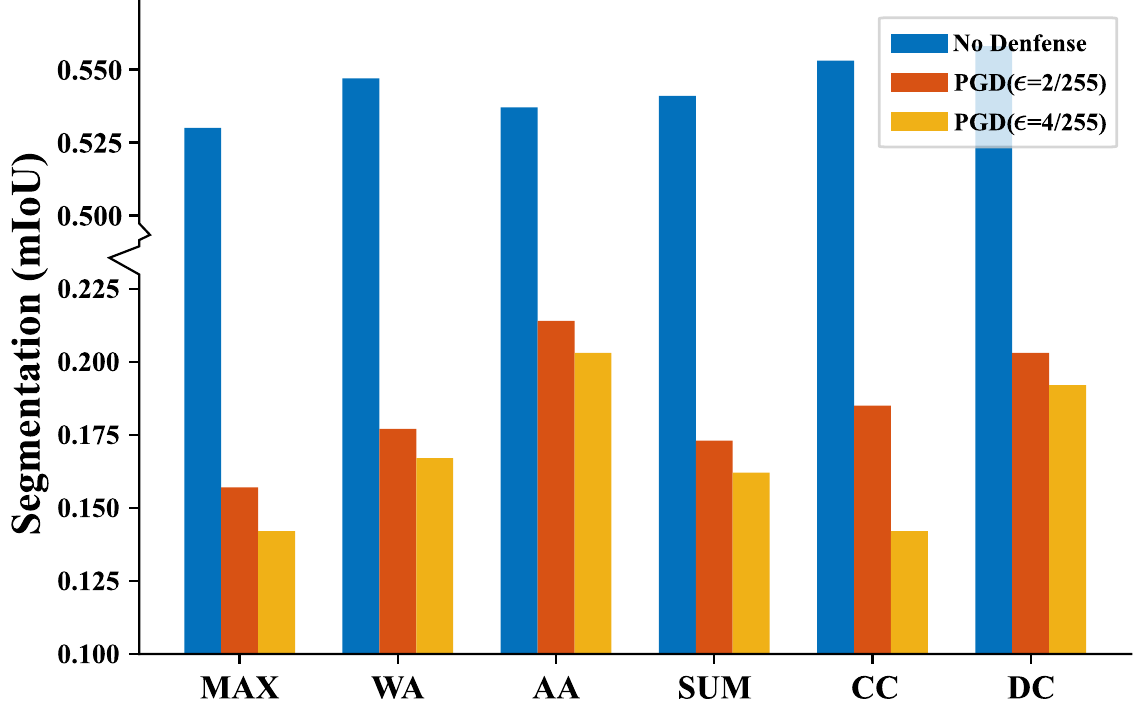}\\
		
		\footnotesize (a) Operations (w/o Attacks) & \footnotesize (b) Operations (PGD $\bm{\bm{\epsilon}} = 2/255$) &\footnotesize (c) Operations (PGD $\bm{\bm{\epsilon}} = 4/255$)  &\footnotesize (d) Fusion Rules
	\end{tabular}
	\vspace{-0.4cm}
	\caption{Robustness analyses of  segmentation results with diverse fusion operations and rules under  adversarial conditions. Subfigure (a), (b) and (c) depict the segmentation performances with diverse fusion operations under various 
		adversarial perturbations. Subfigure (d) plots the segmentation performances of fusion rules.}
	\label{fig:analyze}
\end{figure*}

\subsection{Contributions}
To partially mitigate these issues, by exploring the architectures and parameters learning of image fusion, we make the first attempt to analyze, design and promote the multi-modality perception robustness, to effectively address the adversarial attacks for downstream semantic perception tasks (\textit{e.g.,} segmentation), shown in Fig.~\ref{fig:first}. Introducing the cascaded scheme of image fusion and semantic segmentation,
we thoroughly analyze the performance stability of various fusion models with different operations (\textit{e.g.,} convolutions and attentions) and fusion rules (\textit{e.g.,} maxing selection, average and concatenation) under diverse adversarial perturbations. According to  the instructive finding revealed by these exploratory evaluations, we propose the modality decomposition supernet to further ameliorate the influences of adversarial artifacts and establish a fine-grained search space by discarding the harmful operations. The Harmonized Defence Search (HDS) strategy is proposed to balance the robustness of parameters and efficient architectures. Then we present the Adaptive 
Adversarial Training (AAT) for image fusion under diverse latent attack scenes to learn the robust parameters for the complementary feature extraction and restoration from perturbations.
With the above strategies, we can construct a unified image fusion-based defence framework for multi-modality semantic perception.
Our core contributions can be summarized as:
\begin{itemize}
	\item  To the best of our knowledge, it is the first time to systematically construct the perception-aware image fusion to strengthen the robustness of multi-modality semantic perception, where image fusion not only aggregates typical modality characteristics but also removes the adversarial perturbations for the perception.
	
	\item From the architectural perspective, we enhance the robustness of the fusion network by exploring decomposition-based architectures and constructing a harmonized search strategy to boost the accuracy and robustness jointly.
	
	\item From the learning perspective, we introduce the adaptive adversarial training scheme, to strengthen the consistency of image fusion by optimizing parameters under diverse transferred attack conditions, which effectively facilitates the extraction of typical modalities features.
	
	\item Extensive experimental results demonstrate our scheme empirically realizes the remarkable promotion of both standard accuracy and robustness for  segmentation compared with existing advanced networks and training strategies.  
\end{itemize}
\begin{figure*}[htb]
	\centering
	\includegraphics[width=0.98\textwidth]{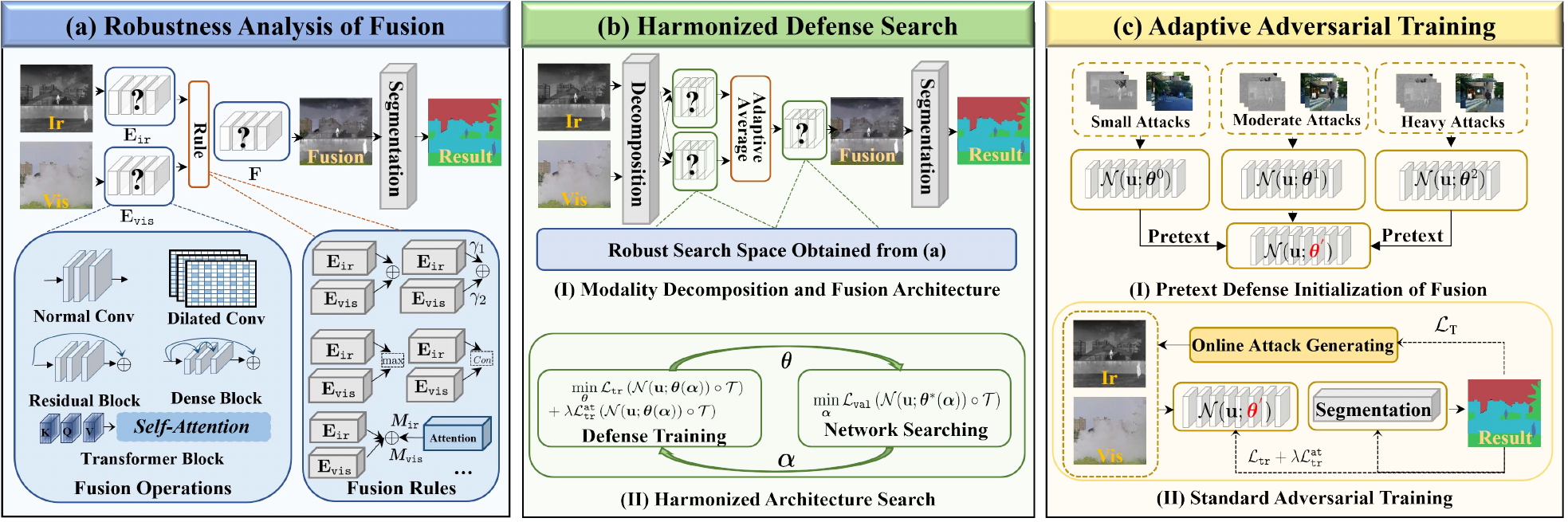}
	\vspace{-0.4cm}
	\caption{{The schematic graph of the proposed framework. We first illustrate the pipeline to evaluate the robustness of diverse fusion operations and rules for segmentation in subfigure (a). In subfigure (b), we propose a decomposition-based network with a harmonized search strategy for robust architecture. Finally, we depict the adaptive adversarial training in subfigure (c).}}
	\label{fig:illustration}
\end{figure*}

\section{Proposed Method}
Specifically, we  formulate the pipeline notations as  $\mathbf{u} = \mathcal{N}(\mathbf{x},\mathbf{y};\bm{\theta})$ and
$\mathbf{z} =\mathcal{T}(\mathbf{u};\bm{\omega})$, where $\mathcal{N}$ and $\mathcal{T}$ denote the networks of fusion and perception task with parameters $\bm{\theta}$ and $\bm{\omega}$ respectively. $\mathbf{x}$, $\mathbf{y}$, $\mathbf{u}$, and $\mathbf{z}$ represent the infrared, visible, fused image and task outputs. 

\subsection{Robustness Analysis of Fusion Network}
Lately, some works investigate robust architectures by the search and manual design for standard vision tasks (\textit{e.g.,} classification)~\cite{su2018robustness,mao2022towards}, but these practices for image fusion to improve the robustness of downstream perception have not been widely studied.
In this part, we investigate the resilient properties of different patterns of fusion networks to adversarial attacks.
We provide detailed analyses of two essential components for multi-modality image fusion, \textit{i.e.,} fusion operation and rules. The baseline model is plotted in Fig.~\ref{fig:illustration} (a), which is widely used in the fusion task~\cite{li2018densefuse,zhao2020didfuse,TarDAL}. 


\textit{Fusion operations.} Operations are the basic modules for fusion networks. We first explore the most utilized operations in existing practices, which includes normal convolutions (denoted as $\mathbf{C}$), dilated convolutions (denoted as $\mathbf{DC}$), residual blocks (denoted as $\mathbf{RB}$), dense blocks (denoted as $\mathbf{DB}$), with kernel size $k\times k, k \in\{3,5,7\}$,
spatial attention (denoted as $\mathbf{SA}$), channel attention (denoted as $\mathbf{CA}$)  and transformer blocks~\cite{wu2022breaking} (denoted as $\mathbf{TB}$). 

\textit{Fusion rules.} The fusion rule is a key part of a multi-modality fusion, aiming to provide a way to integrate the complementary characteristics of modal features. In the following, we summarize five fusion rules of the existing network, which are formulated as:
(1) Maxing Selection (MAX). We formulate the rule as $\mathbf{F} = \max (\mathbf{E}_\mathtt{ir}; \mathbf{E}_\mathtt{vis})$, where $\max$ represents the element-wise maxing operator~\cite{SeaFusion}. We also denote the feature extracted by encoders as $\mathbf{E}_\mathtt{ir}$ and $\mathbf{E}_\mathtt{vis}$ respectively. 
(2) Weighted Average (WA). This strategy needs to manually coordinate  the weights of aggregation~\cite{li2018densefuse,zhao2020didfuse}. Denoted the weights as  $\gamma_{1}$ and $\gamma_{2}$, the rule can be written as  $\mathbf{F} =  \gamma_{1}\mathbf{E}_\mathtt{ir} \oplus \gamma_{2}\mathbf{E}_\mathtt{vis}$. 
(3) Adaptive Average (AA).  By introducing the spatial attention, we generate the mask of source images $\bm{M}_\mathtt{ir}$ and $\bm{M}_\mathtt{vis}$ to introduce the 
adaptive aggregation, \textit{i.e.,}    $\mathbf{F} =  \bm{M}_\mathtt{ir}\mathbf{E}_\mathtt{ir} \oplus \bm{M}_\mathtt{vis}\mathbf{E}_\mathtt{vis}$. 
(4) Summation (SUM). This rule is proposed to aggregate diverse modal features by element-wise addition~\cite{li2018densefuse,prabhakar2017deepfuse}, which can be represented by $\mathbf{F} =  \mathbf{E}_\mathtt{ir} \oplus \mathbf{E}_\mathtt{vis}$.
(5) Channel Concatenation (CC). By concatenating the features at channel dimension, this strategy can be depicted as $\mathbf{F} = Con (\mathbf{E}_\mathtt{ir}; \mathbf{E}_\mathtt{vis})$, where $Con$ represents the concatenation operators.
(6) Direct Combination (DC). Following with works~\cite{U2Fusion2020,TarDAL}, which directly combine the source images as the inputs for network, we define this rule
as  $\mathbf{F} = \mathbf{C}(\mathbf{I}_\mathtt{ir}; \mathbf{I}_\mathtt{vis})$.

To answer the question about segmentation robustness of these fusion operations and rules affected by the attacks, we leverage the hybrid model $\mathcal{N}\circ\mathcal{T}$ to validate them based on the attacks    of downstream semantic perception. We utilize Projected Gradient Descent (PGD)~\cite{madry2017towards}\footnote{ {In this manuscript, we uniformly adopt 5 iterations to generate the adversarial attacks.}} to verify the robust accuracy of these variants, which is formulated as:
\begin{equation}
\mathbf{x}^\mathtt{adv}:= \mathbf{x} + \delta_\mathtt{ir} \leftarrow g(\mathbf{x},\mathbf{y},\bm{\eta},\bm{\epsilon}); \mathbf{y}^\mathtt{adv}:= \mathbf{y} + \delta_\mathtt{vis} \leftarrow g(\mathbf{x},\mathbf{y},\bm{\eta},\bm{\epsilon}),
\label{eq:pgd}
\end{equation}
where $g(\cdot)$ is the PGD attack based on the segmentation loss, $\delta$ represents the adversarial perturbation. $\bm{\eta}$ denotes the step size (setting as $\bm{\epsilon}/4$), and $\bm{\epsilon}$ denotes the perturbation budget regarding to $\ell_{\infty}$-norm. The subscript ``$\mathtt{adv}$'' represents the attacked images. By maximizing the distances between output $\mathbf{z}$ and labels $\mathbf{z}^{*}$, while misleads the model, the objective of adversarial attacks can be formulated as:
\begin{equation}
\delta_\mathtt{ir}, \delta_\mathtt{vis} = \arg\max\limits_{ \delta_\mathtt{ir}, \delta_\mathtt{vis}}\mathcal{L}_\mathtt{T}(\mathcal{N}\circ\mathcal{T}(\mathbf{x}^\mathtt{adv},\mathbf{y}^\mathtt{adv};\bm{\theta},\bm{\omega});\mathbf{z}^{*}),
\label{eq:attack}
\end{equation}
where $\mathcal{L}_\mathtt{T}$ represents the perception-specific loss.
Specifically, we adopt the channel concatenation rule to evaluate the operations. 
We also utilize the same operation (3$\mathbf{RB}$) to validate fusion rules. All numerical results are depicted in Fig.~\ref{fig:analyze} by histograms.

Three critical conclusions can be obtained from the robustness analysis. First, the robustness of fusion can be benefited from the expansion of dilated convolutions rather than kernel sizes.  3$\mathbf{DC}$ has higher robustness compared with 3$\mathbf{C}$. Though better performance under no defense can be realized by enlarging the kernel size, the robustness is substantially decreased. Second, dense connections can effectively boost adversarial attacks compared with a single residual connection. Meanwhile, channel attention can effectively preserve robust features rather than at a spatial level.  Lastly, the adaptive average exhibits the strongest robustness, which stems from the flexible maintenance of modal features, compared to the simple aggregation (\textit{e.g.,} maxing selection and weighted average). 

\vspace{-1.1em}
\subsection{Harmonized Defence Architecture Search}
Following these observations, to realize the robust architecture construction, we first establish a more powerful super-net for image fusion based on the feature decomposition perspective. Then combining the concrete segmentation network, we proposed the harmonized defence searching scheme, as depicted in Fig.~\ref{fig:illustration} (b).

\textbf{Modality decomposition and fusion.} As mentioned above, fusion network $\mathcal{N}$ has two distinct goals, including extracting comprehensive features from diverse modalities and removing the adversarial artifacts. Unfortunately, their components are entangled in each other. Thus,  realizing both goals simultaneously is untoward with the heavy presence of artifacts.
In order to suppress the influences of artifacts from source images, we further introduce the feature decomposition mechanism~\cite{wu2018fast,abs-2211-14461} to disentangle the attacks artifacts and shared information from different modalities. 

Noting that,  some fusion methods also leverage the feature decomposition to extract the backgrounds and textural details from  source images~\cite{liu2021investigating,zhao2021efficient,liu2021searching} directly. Different from these approaches, the proposed feature decomposition is guided by the residual modality features~\cite{li2018robust}, which can be 
expressed by 
\begin{equation}
\mathbf{F}^\mathtt{res}_\mathtt{k}(x) = \max_\mathtt{i \in C} \mathbf{F}_\mathtt{k}^\mathtt{i}(x) -  \min_\mathtt{i \in C} \mathbf{F}_\mathtt{k}^\mathtt{i}(x), \mathtt{k} \in \{\mathtt{ir},\mathtt{vis}\}\label{eq:resfeature}
\end{equation}
The residual feature $\mathbf{F}^\mathtt{res}$ can realize the salient components extraction from source features ($\textit{e.g.,}$ adversarial artifacts). 
In this way, using these features can greatly facilitate the extraction of adversarial artifacts.
In consideration of  performance and robustness, we select four operations for search space, including 3-$\mathbf{DC}$, 7-$\mathbf{RB}$, 3-$\mathbf{DB}$ and $\mathbf{CA}$. The adaptive average is exploited as the fusion rules.

\textbf{Harmonized search strategy.} 
Currently, extensive efforts are devoted to develop the specialized learning strategies. Few attentions are put forward to enhance the robustness from  architecture perspectives, especially utilizing the architecture search.

The proposed 
strategy is based on the differentiable architecture search~\cite{liu2018darts,liu2022optimization,liu2021towards}. By introducing the continuous relaxation $\bm{\alpha}$, the whole optimization objective for search can be formulated as:
\begin{equation}\begin{array}{ll}\label{eq:bilevel}
\min \limits_{\bm{\alpha}} & \mathcal{L}_{\mathtt{val}}\left(\mathcal{N}(\mathbf{u};\bm{\theta}^{*}(\bm{\alpha}))\circ\mathcal{T}\right), \\
\mbox{s.t.} & \min \limits_{\bm{\theta}} \mathcal{L}_{\mathtt{tr}}\left(\mathcal{N}(\mathbf{u};\bm{\theta}(\bm{\alpha}))\circ\mathcal{T}\right) 
+{\lambda} \mathcal{L}_{\mathtt{tr}}^{\mathtt{at}}\left(\mathcal{N}(\mathbf{u};\bm{\theta}(\bm{\alpha}))\circ\mathcal{T}\right),
\end{array}\end{equation}
where $\mathcal{L}_{\mathtt{val}}$, $\mathcal{L}_{\mathtt{tr}}$, and $\mathcal{L}_{\mathtt{tr}}^{\mathtt{at}}$ denote the validation loss, normal training and attacked training loss guided by perception tasks. 

In details, we optimize  the above training objective in a harmonized manner, which can be decoupled into two iterative steps: the robust parameters training for $\bm{\theta}$ and standard performance validation for $\bm{\alpha}$.  Specifically, we warmly start the $\mathcal{N}\circ\mathcal{T}$ jointly to avoid the random attacks at the initial phase. 
Then as for the optimization of $\bm{\theta}$, we utilized mixed data (including normal and adversarial samples)  to conduct the standard adversarial training. In order to strike a balance between the performances and robustness, avoiding the search oscillation, we only utilize the normal samples to optimize the upper-level objective. The core procedure is summarized in Alg.~\ref{alg:framework}.
\begin{algorithm}[htb] 
	\caption{Harmonized Defence Search (HDS).}\label{alg:framework}
	\begin{algorithmic}[1] 
		\REQUIRE Clean datasets with $\{\mathbf{x},\mathbf{y}\}$, perception labels $\mathbf{z}^{*}$,  and other necessary hyper-parameters.
		
		\STATE Pretrain the $\mathcal{N}\circ\mathcal{T}$ for initializing $\bm{\theta}$ and $\bm{\omega}$.
		\WHILE {not converged}
		\STATE \% \emph{Adversarial Training.} 
		\STATE Generating the adversarial sample and setting batch as $\{\mathbf{x}_\mathtt{1},\mathbf{y}_\mathtt{1}, \mathbf{z}_\mathtt{1}^{*};\cdots\mathbf{x}_\mathtt{m}^\mathtt{adv},\mathbf{y}_\mathtt{m}^\mathtt{adv},\mathbf{z}_\mathtt{m}^{*}\}$ and computing the hybrid loss. Updating parameters of $\bm{\theta}$ and $\bm{\omega}$.
		\STATE \% \emph{Architecture Search.} 
		\STATE Updating $\bm{\alpha}$ utilizing the first-order approximation~\cite{liu2022revisiting} and general validation losses on clear dataset.
		\ENDWHILE
		
		\RETURN  $\bm{\alpha}^{*}$.
	\end{algorithmic}
\end{algorithm}

\begin{algorithm}[thb] 
	\caption{Adaptive Adversarial Training (AAT).}\label{alg:training}
	\begin{algorithmic}[1] 
		\REQUIRE Loss function $\mathcal{L}_\mathtt{F}$, and off-line black box-attacked datasets $\mathcal{D}_\mathtt{tr}$ and $\mathcal{D}_\mathtt{val}$.
		
		\STATE \% Pretext defence initialization of fusion.
		\WHILE {not converged}
		\FOR {each attack scene i with several gradient descent} 
		\STATE $\bm{\theta}^\mathtt{i} \leftarrow \bm{\theta}-\nabla_{\bm{\theta}}\mathcal{L}_\mathtt{F}(\mathbf{x}^\mathtt{adv},\mathbf{y}^\mathtt{adv};\bm{\theta};\mathcal{D}_\mathtt{tr}^\mathtt{i}) $. 
		\ENDFOR
		\STATE $\bm{\theta}\leftarrow \bm{\theta}-\nabla_{\bm{\theta}}\sum_{i=0}^{N}\mathcal{L}_\mathtt{F}(\bm{\theta};\mathcal{N}(\bm{\theta}^\mathtt{i});\mathcal{D}_\mathtt{val}^\mathtt{i})$. 
		\ENDWHILE
		\STATE \% Warm start of image fusion.
		\STATE  $\bm{\theta}^{'}\leftarrow \bm{\theta}-\nabla_{\bm{\theta}}\mathcal{L}_\mathtt{F}(\bm{\theta};\mathcal{N}(\mathbf{x}^\mathtt{adv},\mathbf{y}^\mathtt{adv};\bm{\theta});\sum_\mathtt{i}\mathcal{D}_\mathtt{tr}^\mathtt{i})$;
		\STATE \% Adaptive adversarial training.
		\STATE  $\bm{\theta}^{*}$, $\bm{\omega}^{*} =\arg\min_{\bm{\theta},\bm{\omega}}\mathcal{L}_\mathtt{tr}(\mathcal{N}(\bm{\theta}^{'})\circ\mathcal{T}(\bm{\omega}))+\lambda\mathcal{L}_\mathtt{tr}^\mathtt{at}(\mathcal{N}\circ\mathcal{T})$;
		\RETURN  $\bm{\theta}^{*}$, $\bm{\omega}^{*}$.
	\end{algorithmic}
\end{algorithm}
\subsection{Adaptive Adversarial Parameter Training}

The fusion of framework plays two critical roles, \textit{i.e.,} extracting task-desired features from the source modalities and weakening the impacts of adversarial perturbations. The fusion parameters $\bm{\theta}$ should have robust generalization for diverse attacks.
As shown in Fig.~\ref{fig:illustration} (c), we propose the adaptive adversarial training scheme to primarily investigate the parameter robustness of image fusion under diverse attacks for fast adaptation, which can be decoupled into two steps: pretext defence initialization and adaptive adversarial training. 
The whole optimization can be summarized in Alg.~\ref{alg:training}. Firstly, inspired by the transfer-based attacks~\cite{dong2019evading,xie2019improving}, which transfer the adversarial samples of the source model into the target model, we set up three different levels of perturbations. To enhance the relevance  of  attacks, we utilize the pretrained $\mathcal{N}\circ\mathcal{T}$ under standard adversarial training to composite the source model, which is composited by 3-$\mathbf{DB}$ and integrated by the concatenation rule. In concrete, we generate different degrees of attacks in an offline manner, composited by small, moderate, and heavy attacks. The objective of the whole pretext defence can be formulated as:
\begin{equation}
\bm{\theta}^{'} = \arg\min_{\bm{\theta}}\sum\limits_{i=0}^{N}\mathcal{L}_\mathtt{F}(\bm{\theta};\mathcal{N}(\mathbf{x}^\mathtt{adv},\mathbf{y}^\mathtt{adv};\bm{\theta}^\mathtt{i})),\label{eq:pretext}
\end{equation}
where $N$ denotes the categories of attacks and $\mathcal{L}_\mathtt{F}$ represents the designed fusion loss. The details are elaborated on in Section 3.1. A hierarchical strategy is utilized for solving this objective~\cite{finn2017model}. We first obtain the $\bm{\theta}^\mathtt{i}$ on specific attacks by several gradient steps from the initialized $\bm{\theta}$ (line 4 in Alg.~\ref{alg:training}). Then we endow the generalization of fusion based on the optimization of $\bm{\theta}$ from diverse $\bm{\theta}^\mathtt{i}$.
Finally, we assign the well-initialized  $\bm{\theta}^{'}$ into the fusion  and conduct the standard adversarial training jointly.
\begin{table*}[thb]
	\centering
	
	\caption{ Quantitative results of semantic segmentation  compared with different methods on the {MFNet} dataset.}~\label{tab:seg}
	
	\renewcommand{\arraystretch}{1.1}
	\vspace{-0.4cm}
	\setlength{\tabcolsep}{0.25mm}{
		\begin{tabular}{c|cccccc|cccccc|cccccc}
			\hline
			\multirow{2}{*}{Methods} & \multicolumn{6}{c|}{Normal Training}                                                                                                                         & \multicolumn{6}{c|}{
				PGD ($\bm{\bm{\epsilon}} = 4/255$) }                                                                                                                                               & \multicolumn{6}{c}{PGD ($\bm{\bm{\epsilon}} = 8/255$)}                                                                                                                      \\ \cline{2-19} 
			& \multicolumn{1}{c|}{\cellcolor{gray!20} Car} & \multicolumn{1}{c|}{\cellcolor{gray!20}Person} & \multicolumn{1}{c|}{\cellcolor{gray!20}Bike} & \multicolumn{1}{c|}{\cellcolor{gray!20}Cone} & \multicolumn{1}{c|}{\cellcolor{gray!20}Bump} & \cellcolor{gray!20} mIOU $\uparrow$ & \multicolumn{1}{c|}{\cellcolor{gray!20} Car} & \multicolumn{1}{c|}{\cellcolor{gray!20} Person} & \multicolumn{1}{c|}{\cellcolor{gray!20} Bike} & \multicolumn{1}{c|}{\cellcolor{gray!20} Cone} & \multicolumn{1}{c|}{\cellcolor{gray!20}Bump} & \multicolumn{1}{c|}{\cellcolor{gray!20} mIOU $\uparrow$ } & \multicolumn{1}{c|}{\cellcolor{gray!20} Car} & \multicolumn{1}{c|}{\cellcolor{gray!20}Person} & \multicolumn{1}{c|}{\cellcolor{gray!20} Bike} & \multicolumn{1}{c|}{\cellcolor{gray!20} Cone} & \multicolumn{1}{c|}{\cellcolor{gray!20} Bump} & \cellcolor{gray!20} mIOU $\uparrow$  \\ \hline
			U2Fusion	& \multicolumn{1}{c|}{\textcolor{blue}{\textbf{0.827}} }    & \multicolumn{1}{c|}{0.641}       & \multicolumn{1}{c|}{\textcolor{red}{\textbf{0.610}}}     & \multicolumn{1}{c|}{0.463}     & \multicolumn{1}{c|}{\textcolor{blue}{\textbf{0.499}} }     &\textcolor{blue}{\textbf{0.514}}    & \multicolumn{1}{c|}{0.478}    & \multicolumn{1}{c|}{0.316}       & \multicolumn{1}{c|}{0.336}     & \multicolumn{1}{c|}{0.104}     & \multicolumn{1}{c|}{0.000}     &  0.254                         & \multicolumn{1}{c|}{0.306}    & \multicolumn{1}{c|}{0.177}       & \multicolumn{1}{c|}{0.238}     & \multicolumn{1}{c|}{0.053}     & \multicolumn{1}{c|}{0.000}     &0.193      \\ \hline
			DIDFuse	& \multicolumn{1}{c|}{0.780}    & \multicolumn{1}{c|}{0.596}       & \multicolumn{1}{c|}{0.534}     & \multicolumn{1}{c|}{0.347}     & \multicolumn{1}{c|}{0.200}     & 0.433     & \multicolumn{1}{c|}{0.615}    & \multicolumn{1}{c|}{0.465}       & \multicolumn{1}{c|}{0.319}     & \multicolumn{1}{c|}{0.183}     & \multicolumn{1}{c|}{0.000}     &  \textcolor{blue}{\textbf{0.313}}                         & \multicolumn{1}{c|}{\textcolor{blue}{\textbf{0.513}} }    & \multicolumn{1}{c|}{0.312}       & \multicolumn{1}{c|}{0.201}     & \multicolumn{1}{c|}{0.120}     & \multicolumn{1}{c|}{0.000}     &   \textcolor{blue}{\textbf{0.252}}  \\ \hline
		
			SeaFusion	& \multicolumn{1}{c|}{0.822}    & \multicolumn{1}{c|}{\textcolor{blue}{\textbf{0.677}} }       & \multicolumn{1}{c|}{0.597}     & \multicolumn{1}{c|}{0.413}     & \multicolumn{1}{c|}{0.272}     &  0.481    & \multicolumn{1}{c|}{0.548}    & \multicolumn{1}{c|}{\textcolor{blue}{\textbf{0.556}} }       & \multicolumn{1}{c|}{0.206}     & \multicolumn{1}{c|}{0.171}     & \multicolumn{1}{c|}{0.000}     & 0.286                          & \multicolumn{1}{c|}{0.344}    & \multicolumn{1}{c|}{\textcolor{blue}{\textbf{0.344}}}       & \multicolumn{1}{c|}{\textcolor{blue}{\textbf{0.426}}}     & \multicolumn{1}{c|}{0.072}     & \multicolumn{1}{c|}{0.000}     &  0.211    \\ \hline
			SDNet	& \multicolumn{1}{c|}{0.806}    & \multicolumn{1}{c|}{0.658}      & \multicolumn{1}{c|}{0.540}     & \multicolumn{1}{c|}{\textcolor{blue}{\textbf{0.563}} }    & \multicolumn{1}{c|}{0.431}    &  {0.470}    & \multicolumn{1}{c|}{0.594}    & \multicolumn{1}{c|}{0.433}       & \multicolumn{1}{c|}{0.197}     & \multicolumn{1}{c|}{0.105}     & \multicolumn{1}{c|}{0.000}     &    0.277                       & \multicolumn{1}{c|}{0.492}    & \multicolumn{1}{c|}{0.310}       & \multicolumn{1}{c|}{0.121}     & \multicolumn{1}{c|}{0.090}     & \multicolumn{1}{c|}{0.000}     &  0.234    \\ \hline
				ReCoNet	& \multicolumn{1}{c|}{0.793}    & \multicolumn{1}{c|}{0.609}       & \multicolumn{1}{c|}{0.582}     & \multicolumn{1}{c|}{0.367}     & \multicolumn{1}{c|}{0.135}     &  0.446    & \multicolumn{1}{c|}{\textcolor{blue}{\textbf{0.627}} }    & \multicolumn{1}{c|}{0.363}       & \multicolumn{1}{c|}{0.291}     & \multicolumn{1}{c|}{0.200}     & \multicolumn{1}{c|}{0.000}     &   0.295                        & \multicolumn{1}{c|}{0.494}    & \multicolumn{1}{c|}{0.187}       & \multicolumn{1}{c|}{0.201}     & \multicolumn{1}{c|}{0.134}     & \multicolumn{1}{c|}{0.000}     &  0.231    \\ \hline
			
			UMFusion	& \multicolumn{1}{c|}{0.814}    & \multicolumn{1}{c|}{0.625}       & \multicolumn{1}{c|}{0.601}     & \multicolumn{1}{c|}{0.396}     & \multicolumn{1}{c|}{0.468}     &0.491      & \multicolumn{1}{c|}{0.608}    & \multicolumn{1}{c|}{0.404}       & \multicolumn{1}{c|}{\textcolor{blue}{\textbf{0.363}} }     & \multicolumn{1}{c|}{\textcolor{blue}{\textbf{0.257}} }     & \multicolumn{1}{c|}{\textcolor{blue}{\textbf{0.015}} }     &   0.311                        & \multicolumn{1}{c|}{0.453}    & \multicolumn{1}{c|}{0.199}       & \multicolumn{1}{c|}{0.226}     & \multicolumn{1}{c|}{\textcolor{blue}{\textbf{0.163}}}     & \multicolumn{1}{c|}{0.000}     &   0.233   \\ \hline
			TarDAL	& \multicolumn{1}{c|}{0.795}    & \multicolumn{1}{c|}{0.673}       & \multicolumn{1}{c|}{0.599}     & \multicolumn{1}{c|}{0.356}     & \multicolumn{1}{c|}{0.404}     &  0.482    & \multicolumn{1}{c|}{0.538}    & \multicolumn{1}{c|}{0.249}       & \multicolumn{1}{c|}{0.226}     & \multicolumn{1}{c|}{0.207}     & \multicolumn{1}{c|}{0.022}     & 0.264                          & \multicolumn{1}{c|}{0.431}    & \multicolumn{1}{c|}{0.164}       & \multicolumn{1}{c|}{0.138}     & \multicolumn{1}{c|}{0.126}     & \multicolumn{1}{c|}{0.000}     & 0.213     \\ \hline
			Ours	& \multicolumn{1}{c|}{\textcolor{red}{\textbf{0.881}} }    & \multicolumn{1}{c|}{\textcolor{red}{\textbf{0.724}} }       & \multicolumn{1}{c|}{\textcolor{blue}{\textbf{0.608}} }     & \multicolumn{1}{c|}{\textcolor{red}{\textbf{0.560}} }     & \multicolumn{1}{c|}{\textcolor{red}{\textbf{0.572}} }     & \textcolor{red}{\textbf{0.565}}       	& \multicolumn{1}{c|}{\textcolor{red}{\textbf{0.643}} }    & \multicolumn{1}{c|}{\textcolor{red}{\textbf{0.563}} }       & \multicolumn{1}{c|}{\textcolor{red}{\textbf{0.437}} }     & \multicolumn{1}{c|}{\textcolor{red}{\textbf{0.278}} }     & \multicolumn{1}{c|}{\textcolor{red}{\textbf{0.041}} }     &      \textcolor{red}{\textbf{0.361}}             
			& \multicolumn{1}{c|}{\textcolor{red}{\textbf{0.603}} }    & \multicolumn{1}{c|}{\textcolor{red}{\textbf{0.466}} }       & \multicolumn{1}{c|}{\textcolor{blue}{\textbf{0.329}} }     & \multicolumn{1}{c|}{\textcolor{red}{\textbf{0.253}}}     & \multicolumn{1}{c|}{0.000}     &  \textcolor{red}{\textbf{0.314}}    \\ \hline
		\end{tabular}
		
	}
\end{table*}
\begin{figure*}[thb]
	\centering
	\includegraphics[width=0.99\textwidth]{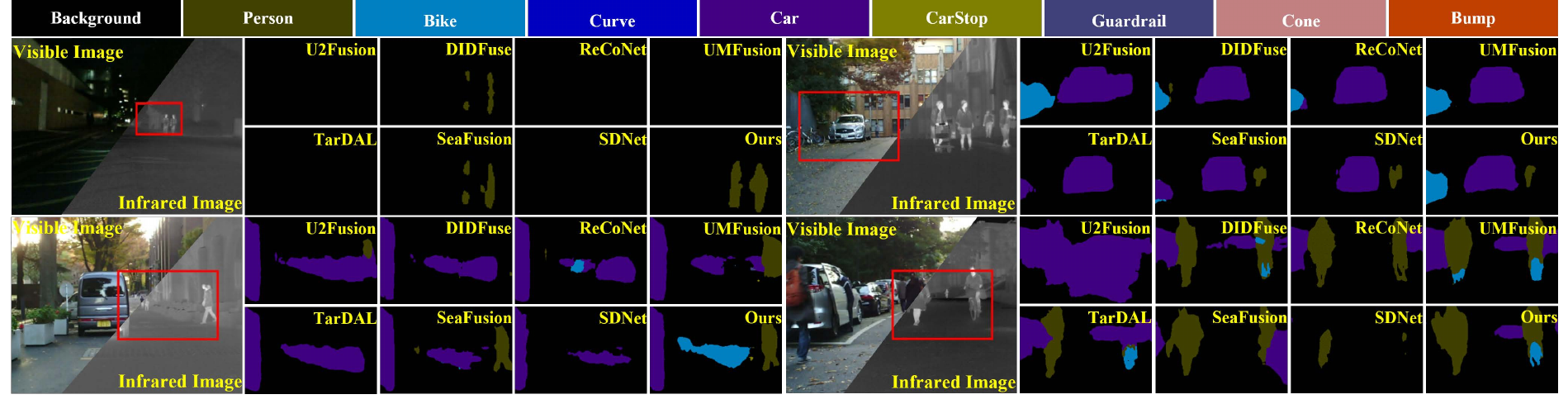}
	\vspace{-0.4cm}
	\caption{{Comparison with advanced image fusion methods for semantic segmentation under diverse attack perturbations. The instances of  each  rows are attacked by PGD with $\bm{\bm{\epsilon}} = 4/255$ and  $8/255$ respectively.}}
	\label{fig:segcomp}
\end{figure*}
\begin{table*}[htb]
	\centering
	\caption{ Quantitative  results of object detection on the M3FD  dataset under diverse adversarial perturbations.}~\label{tab:detec}
	\vspace{-0.4cm}
	\renewcommand{\arraystretch}{1.1}
	\setlength{\tabcolsep}{0.5mm}{
		
		\begin{tabular}{c|ccccccc|cccccc|ccc}
			\hline
			\multirow{2}{*}{Methods} & \multicolumn{1}{c|}{Clean} &\multicolumn{6}{c|}{PGD  ($\bm{\bm{\epsilon}}=1/255$)}                                                                                                                          &
			\multicolumn{6}{c|}{PGD ($\bm{\bm{\epsilon}}=4/255$)}                                                                          & \multicolumn{3}{c}{Efficiency Analysis}                 \\ \cline{2-17} 
			& \cellcolor{gray!20} mAP $\uparrow$  &\cellcolor{gray!20} {Car} & \cellcolor{gray!20}{Bus} & \cellcolor{gray!20}{Motor} & \cellcolor{gray!20}{Truck} & \cellcolor{gray!20}{People} & \cellcolor{gray!20} mAP $\uparrow$&\cellcolor{gray!20} {Car} & \cellcolor{gray!20}{Bus} & \cellcolor{gray!20} {Motor}   & \cellcolor{gray!20}{Truck} & \cellcolor{gray!20}{People} & \cellcolor{gray!20} mAP $\uparrow$  &  \cellcolor{gray!20}{Size(M)$\downarrow$}& \cellcolor{gray!20}{FLOPs(G)$\downarrow$}& \cellcolor{gray!20}{Time(S)$\downarrow$}\\ \hline
			U2Fusion	 & \multicolumn{1}{c|}{0.296} & \multicolumn{1}{c|}{0.490}  & \multicolumn{1}{c|}{0.184}   & \multicolumn{1}{c|}{0.068}   & \multicolumn{1}{c|}{0.093} & \multicolumn{1}{c|}{0.466} &0.229& \multicolumn{1}{c|}{0.371} & \multicolumn{1}{c|}{0.144} & \multicolumn{1}{c|}{0.021} & \multicolumn{1}{c|}{\textcolor{blue}{\textbf{0.042}}}  & \multicolumn{1}{c|}{\textcolor{red}{\textbf{0.394}}} & \multicolumn{1}{c|}{\textcolor{blue}{\textbf{0.172}}} & \multicolumn{1}{c|}{0.659}& \multicolumn{1}{c|}{366.34}& \multicolumn{1}{c}{0.123}\\ \hline
			
			DIDFuse	 & \multicolumn{1}{c|}{0.166} & \multicolumn{1}{c|}{0.465}  & \multicolumn{1}{c|}{0.000}   & \multicolumn{1}{c|}{0.000}   & \multicolumn{1}{c|}{0.000} & \multicolumn{1}{c|}{0.434} & 0.150  & \multicolumn{1}{c|}{0.343} & \multicolumn{1}{c|}{0.000} & \multicolumn{1}{c|}{0.000} & \multicolumn{1}{c|}{0.000}  & \multicolumn{1}{c|}{0.341} & \multicolumn{1}{c|}{0.114} & \multicolumn{1}{c|}{0.373}& \multicolumn{1}{c|}{103.56}& \multicolumn{1}{c}{0.118}\\ \hline

			SDNet	 & \multicolumn{1}{c|}{\textcolor{blue}{\textbf{0.420}}} & \multicolumn{1}{c|}{0.497}  & \multicolumn{1}{c|}{\textcolor{blue}{\textbf{0.248}}}   & \multicolumn{1}{c|}{\textcolor{blue}{\textbf{0.114}}}   & \multicolumn{1}{c|}{\textcolor{red}{\textbf{0.112}}} & \multicolumn{1}{c|}{0.409} &\textcolor{blue}{\textbf{0.252}}  & \multicolumn{1}{c|}{0.274} & \multicolumn{1}{c|}{\textcolor{blue}{\textbf{0.195}}} & \multicolumn{1}{c|}{\textcolor{blue}{\textbf{0.074}}} & \multicolumn{1}{c|}{0.024}  & \multicolumn{1}{c|}{0.223} & \multicolumn{1}{c|}{0.138} & \multicolumn{1}{c|}{\textcolor{red}{\textbf{0.067}}}& \multicolumn{1}{c|}{\textcolor{blue}{\textbf{22.89}}}& \multicolumn{1}{c}{0.041}\\ \hline
						
			ReCoNet	 & \multicolumn{1}{c|}{0.253} & \multicolumn{1}{c|}{0.420}  & \multicolumn{1}{c|}{0.144}   & \multicolumn{1}{c|}{0.029}   & \multicolumn{1}{c|}{0.081} & \multicolumn{1}{c|}{0.335} & {0.181}  & \multicolumn{1}{c|}{0.285} & \multicolumn{1}{c|}{0.123} & \multicolumn{1}{c|}{0.014} & \multicolumn{1}{c|}{0.024}  & \multicolumn{1}{c|}{0.222} & \multicolumn{1}{c|}{0.119} & \multicolumn{1}{c|}{\textcolor{blue}{\textbf{0.209}}}& \multicolumn{1}{c|}{\textcolor{red}{\textbf{12.54}}}& \multicolumn{1}{c}{0.051}\\ \hline
			
			UMFusion	 & \multicolumn{1}{c|}{0.207} & \multicolumn{1}{c|}{\textcolor{blue}{\textbf{0.498}}}  & \multicolumn{1}{c|}{0.110}   & \multicolumn{1}{c|}{0.000}   & \multicolumn{1}{c|}{0.008} & \multicolumn{1}{c|}{0.460} & 0.192  & \multicolumn{1}{c|}{\textcolor{blue}{\textbf{0.378}}} & \multicolumn{1}{c|}{0.107} & \multicolumn{1}{c|}{0.000} & \multicolumn{1}{c|}{0.003}  & \multicolumn{1}{c|}{0.379} & \multicolumn{1}{c|}{0.153} & \multicolumn{1}{c|}{0.629}& \multicolumn{1}{c|}{174.69}& \multicolumn{1}{c}{0.044}\\ \hline
			
			TarDAL	 & \multicolumn{1}{c|}{0.253} & \multicolumn{1}{c|}{0.482}  & \multicolumn{1}{c|}{0.194}   & \multicolumn{1}{c|}{0.000}   & \multicolumn{1}{c|}{0.053} & \multicolumn{1}{c|}{\textcolor{blue}{\textbf{0.472}}} & 0.212  & \multicolumn{1}{c|}{0.339} & \multicolumn{1}{c|}{0.140} & \multicolumn{1}{c|}{0.000} & \multicolumn{1}{c|}{0.032}  & \multicolumn{1}{c|}{0.378} & \multicolumn{1}{c|}{0.156} & \multicolumn{1}{c|}{0.297}& \multicolumn{1}{c|}{82.37}& \multicolumn{1}{c}{\textcolor{red}{\textbf{0.001}}} \\ \hline
			Ours	 & \multicolumn{1}{c|}{\textcolor{red}{\textbf{0.483}}} & \multicolumn{1}{c|}{\textcolor{red}{\textbf{0.531}}}  & \multicolumn{1}{c|}{\textcolor{red}{\textbf{0.314}}}   & \multicolumn{1}{c|}{\textcolor{red}{\textbf{0.152}}}   & \multicolumn{1}{c|}{\textcolor{blue}{\textbf{0.097}}} & \multicolumn{1}{c|}{\textcolor{red}{\textbf{0.484}}} & \textcolor{red}{\textbf{0.283}}  & \multicolumn{1}{c|}{\textcolor{red}{\textbf{0.404}}} & \multicolumn{1}{c|}{\textcolor{red}{\textbf{0.260}}} & \multicolumn{1}{c|}{\textcolor{red}{\textbf{0.115}}} & \multicolumn{1}{c|}{\textcolor{red}{\textbf{0.038}}}  & \multicolumn{1}{c|}{\textcolor{blue}{\textbf{0.389}}} & \multicolumn{1}{c|}{\textcolor{red}{\textbf{0.214}}} & \multicolumn{1}{c|}{0.260} & \multicolumn{1}{c|}{72.18}& \multicolumn{1}{c}{\textcolor{blue}{\textbf{0.007}}} \\ \hline
			
	\end{tabular}}
	
\end{table*}

We argue that this is the first attempt to uniformly consider adversarial training for multi-modality semantic segmentation. In the literature, there are a series of adversarial defence methods (\textit{e.g.,} preprocessing for removing noises~\cite{liao2018defense,xie2019feature},  robust architecture designing~\cite{mao2022towards,guo2020meets} and specialized training strategies~\cite{xie2019improving}). 
The most important difference is that image fusion cannot be simply considered as the ``denoiser'', where the fusion network plays a critical role to aggregate 
the complementary characteristics from attacked modalities. 
Furthermore, our scheme can automatically construct the fusion module without changing the following perception networks. 
Instead of introducing auxiliary mechanisms~\cite{xie2019improving} and loss functions~\cite{chen2021class},  we furthermore investigate the distribution of attacks from diverse scenes for robust feature extraction with strong generalization ability. 
In conclusion, the proposed scheme is comprehensive enough to design strong structure and adaptive training strategy for diverse adversarial conditions.

\begin{figure*}[htb]
	\centering
	\setlength{\tabcolsep}{1pt}
	\begin{tabular}{cccccc}
		\includegraphics[width=0.163\textwidth]{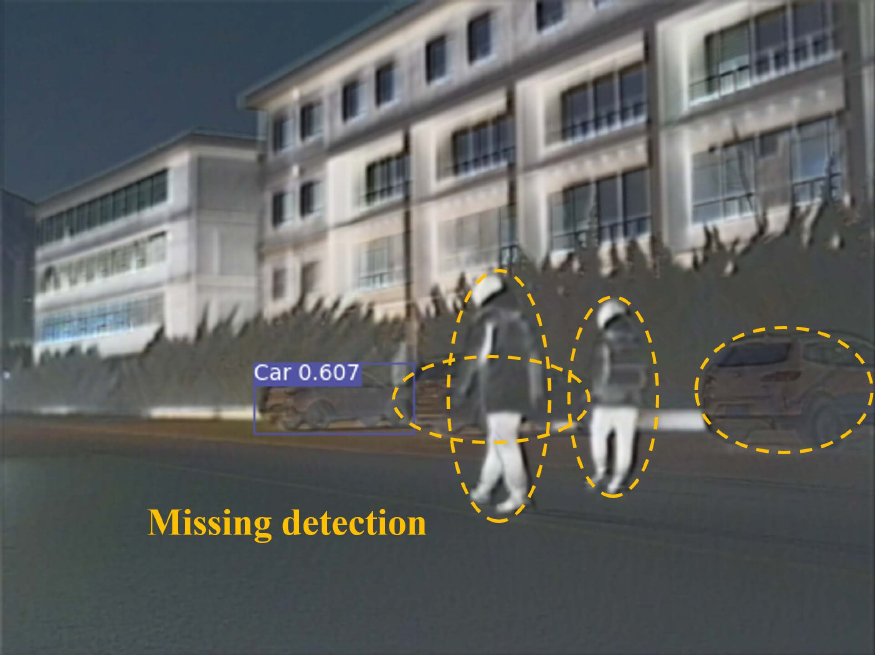}
		&\includegraphics[width=0.163\textwidth]{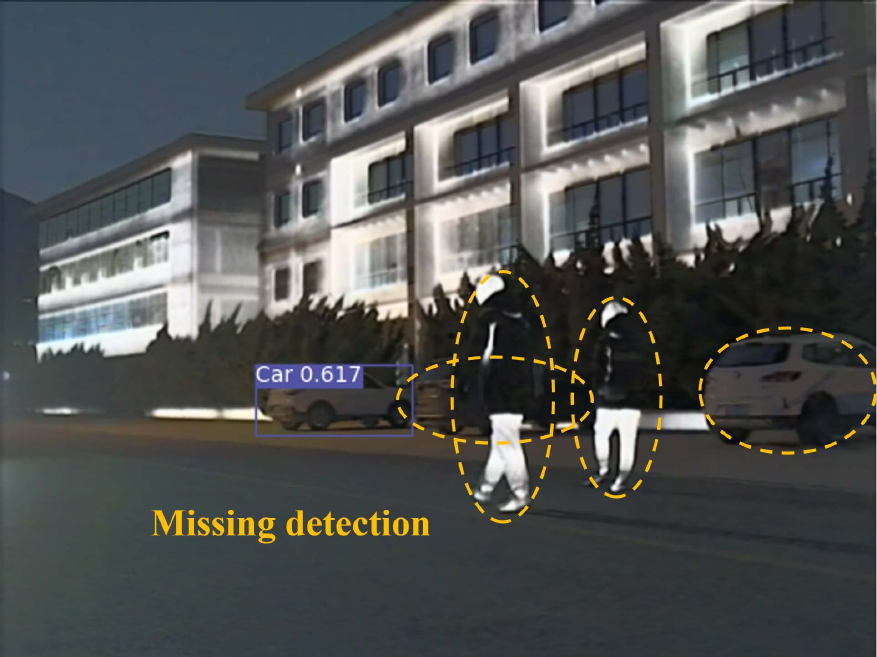}
		&\includegraphics[width=0.163\textwidth]{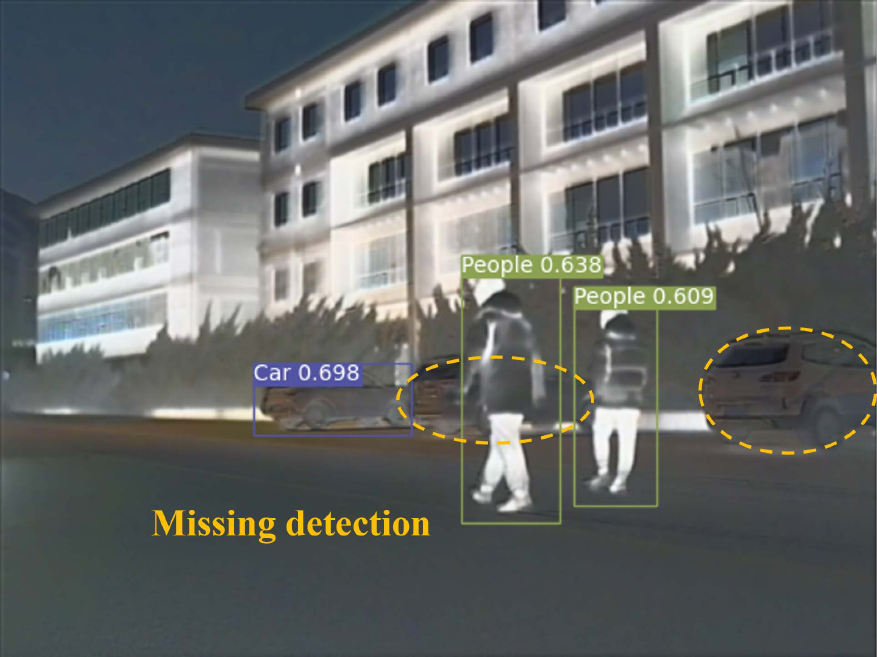}
		&\includegraphics[width=0.163\textwidth]{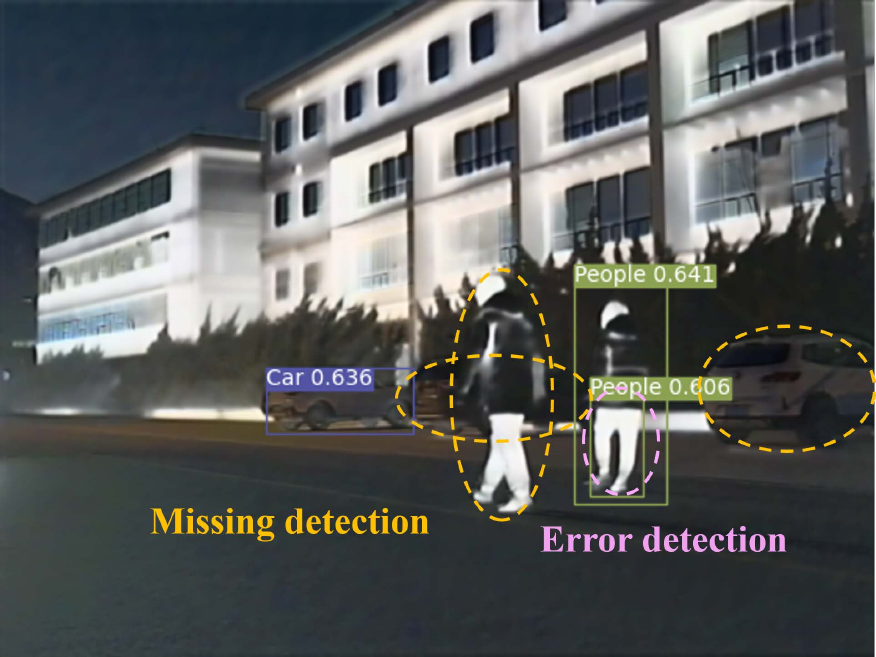}
		&\includegraphics[width=0.163\textwidth]{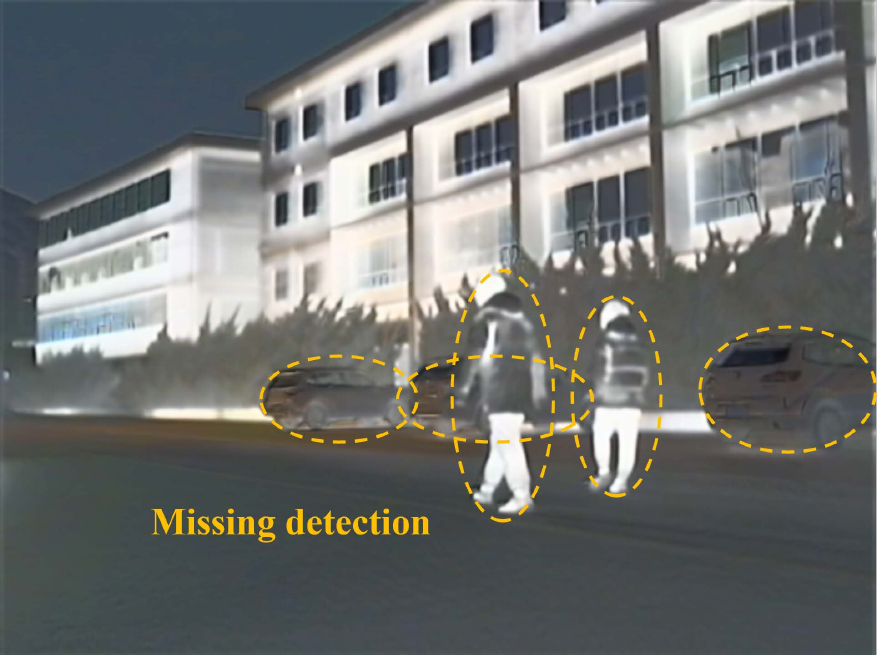}
		&\includegraphics[width=0.163\textwidth]{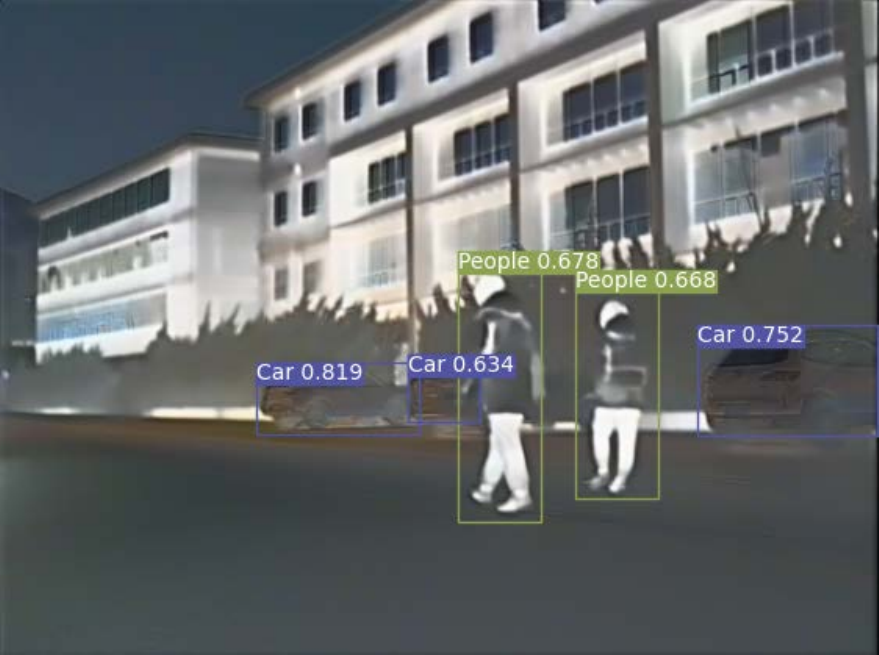}\\
		\includegraphics[width=0.163\textwidth]{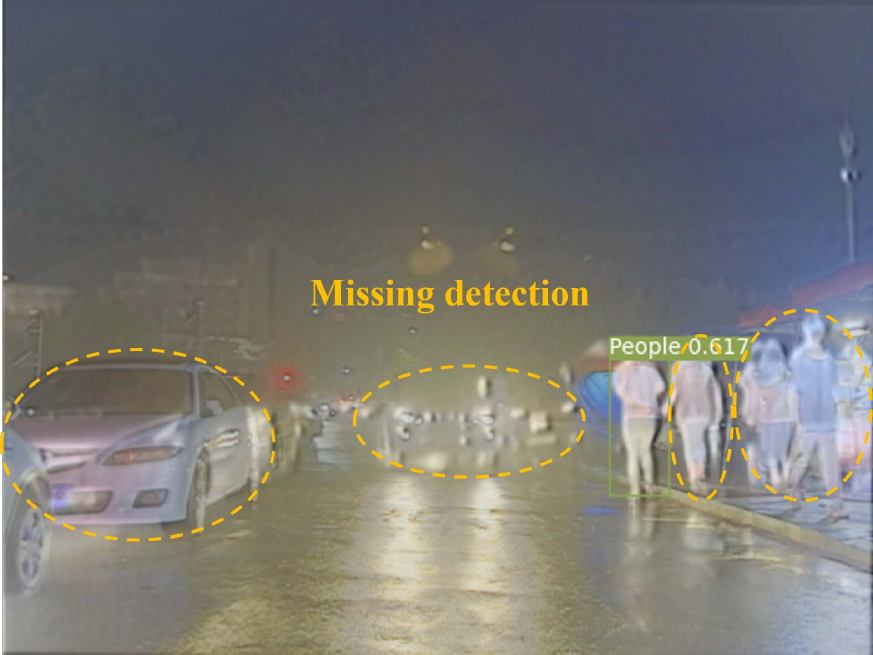}
		&\includegraphics[width=0.163\textwidth]{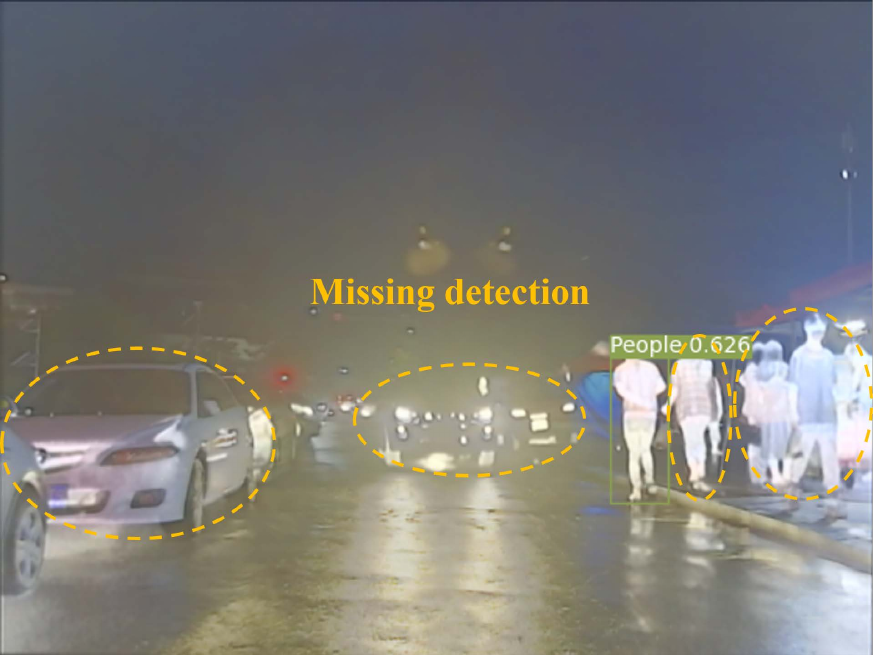}
		&\includegraphics[width=0.163\textwidth]{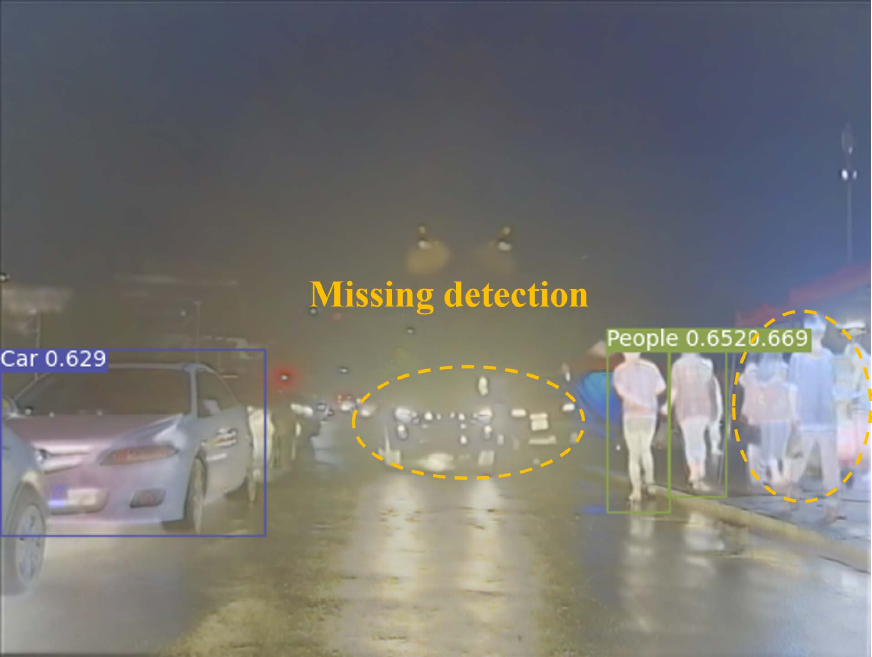}
		&\includegraphics[width=0.163\textwidth]{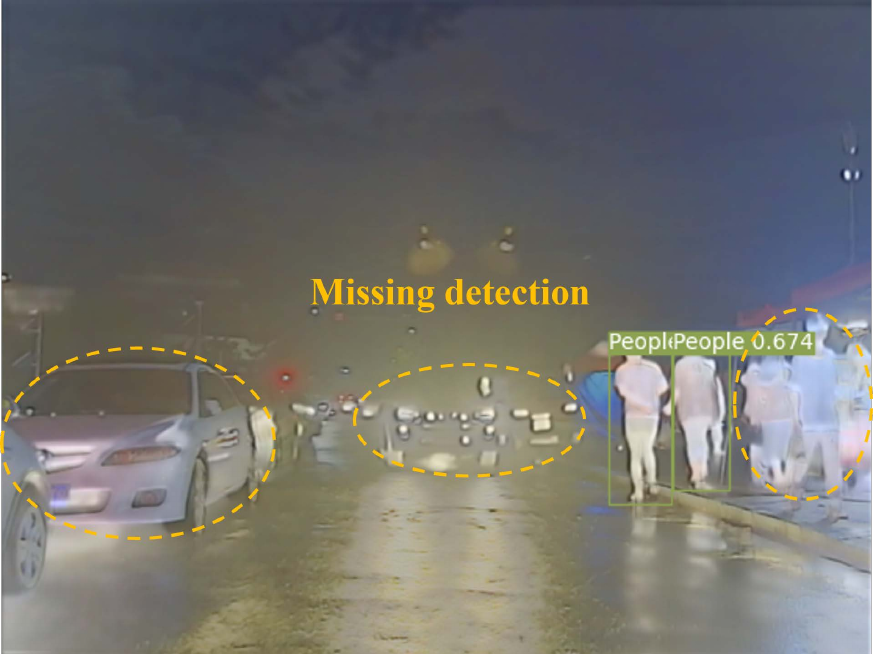}
		&\includegraphics[width=0.163\textwidth]{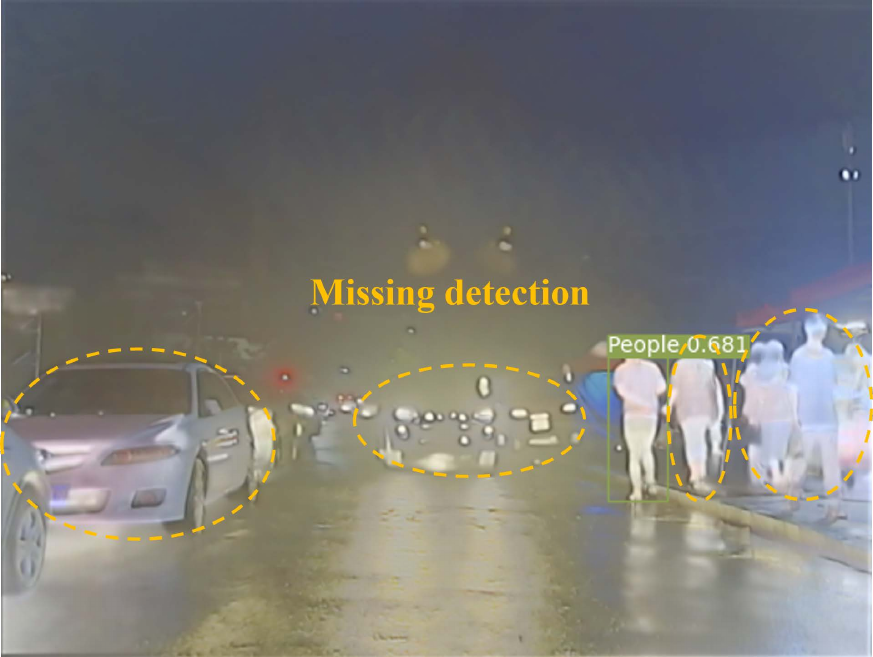}
		&\includegraphics[width=0.163\textwidth]{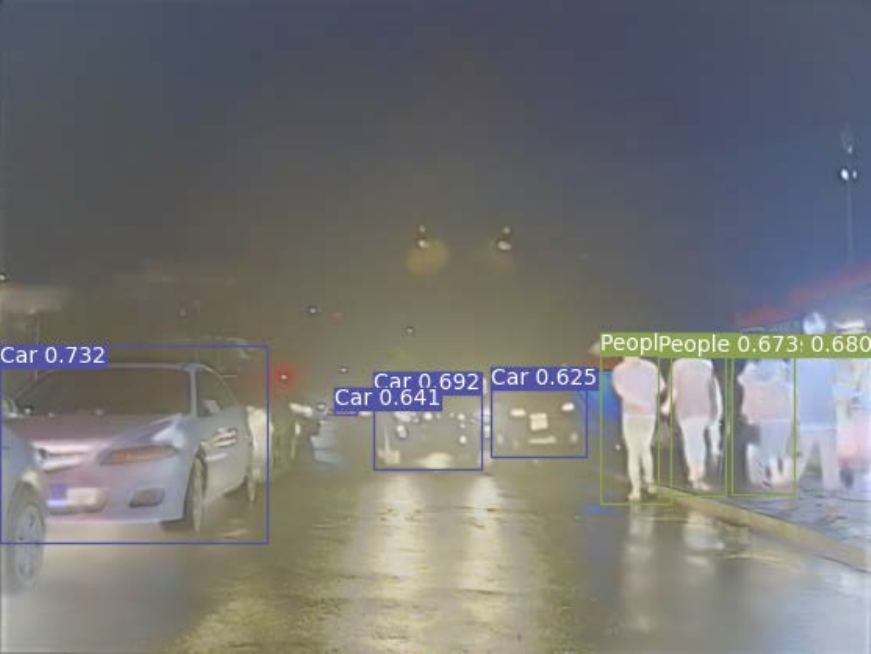}\\
		\footnotesize ReCoNet&\footnotesize UMFusion&\footnotesize DIDFuse&\footnotesize U2Fusion&\footnotesize TarDAL& \footnotesize Ours
		\\
	\end{tabular}
	\vspace{-1em}
	\caption{Visual comparisons of object detection with existing advanced fusion-based detectors  under the similar adversarial attacks (PGD with $\bm{\bm{\epsilon}} = 1/255$) and severe degradations. We mark undetected objects with orange dotted ellipses.}
	\label{fig:detec}
\end{figure*}
\begin{figure*}[htb]
	\centering
	\setlength{\tabcolsep}{1pt}
	\begin{tabular}{ccccccc}
		\includegraphics[width=0.135\textwidth]{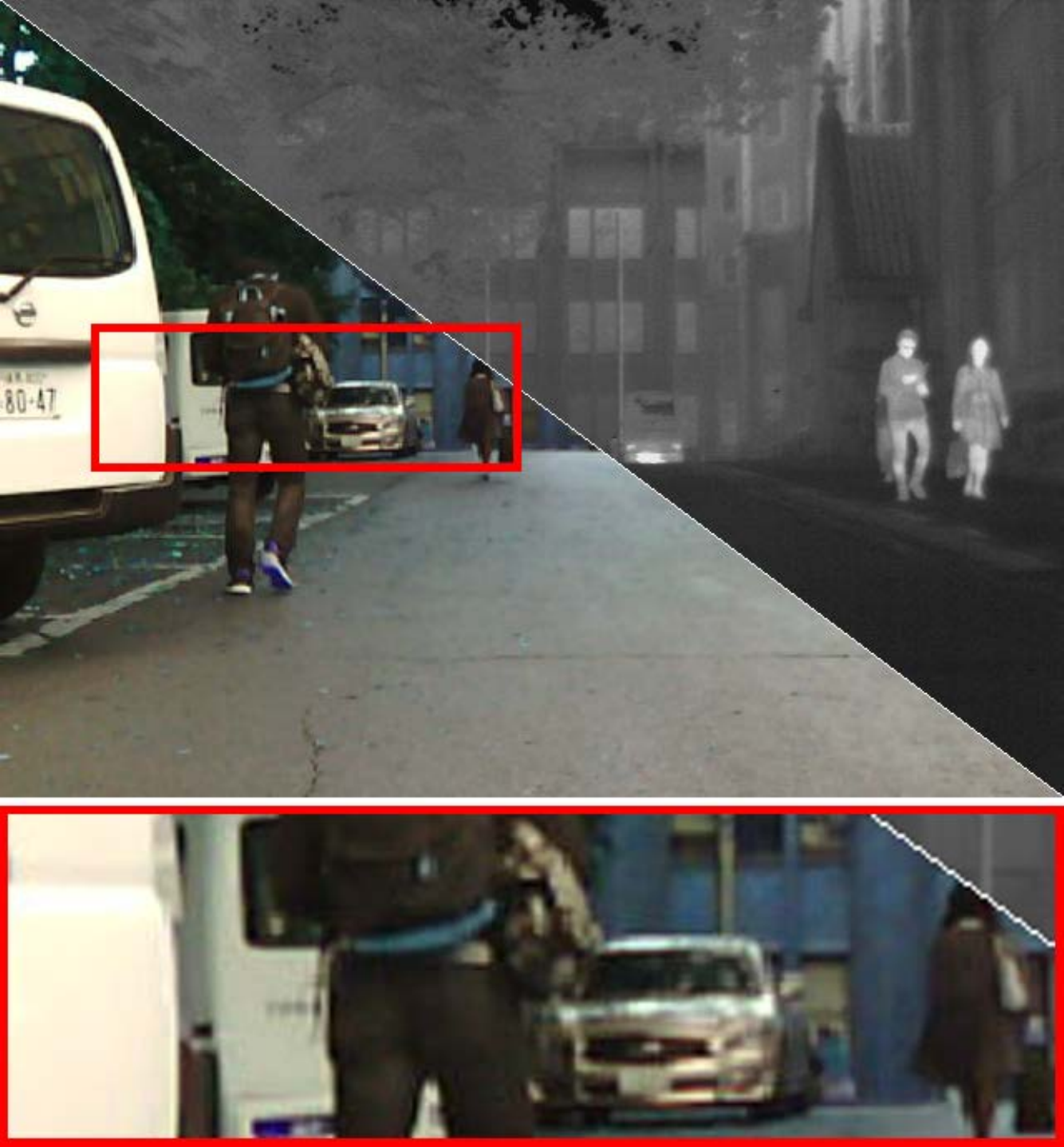}
		&\includegraphics[width=0.135\textwidth]{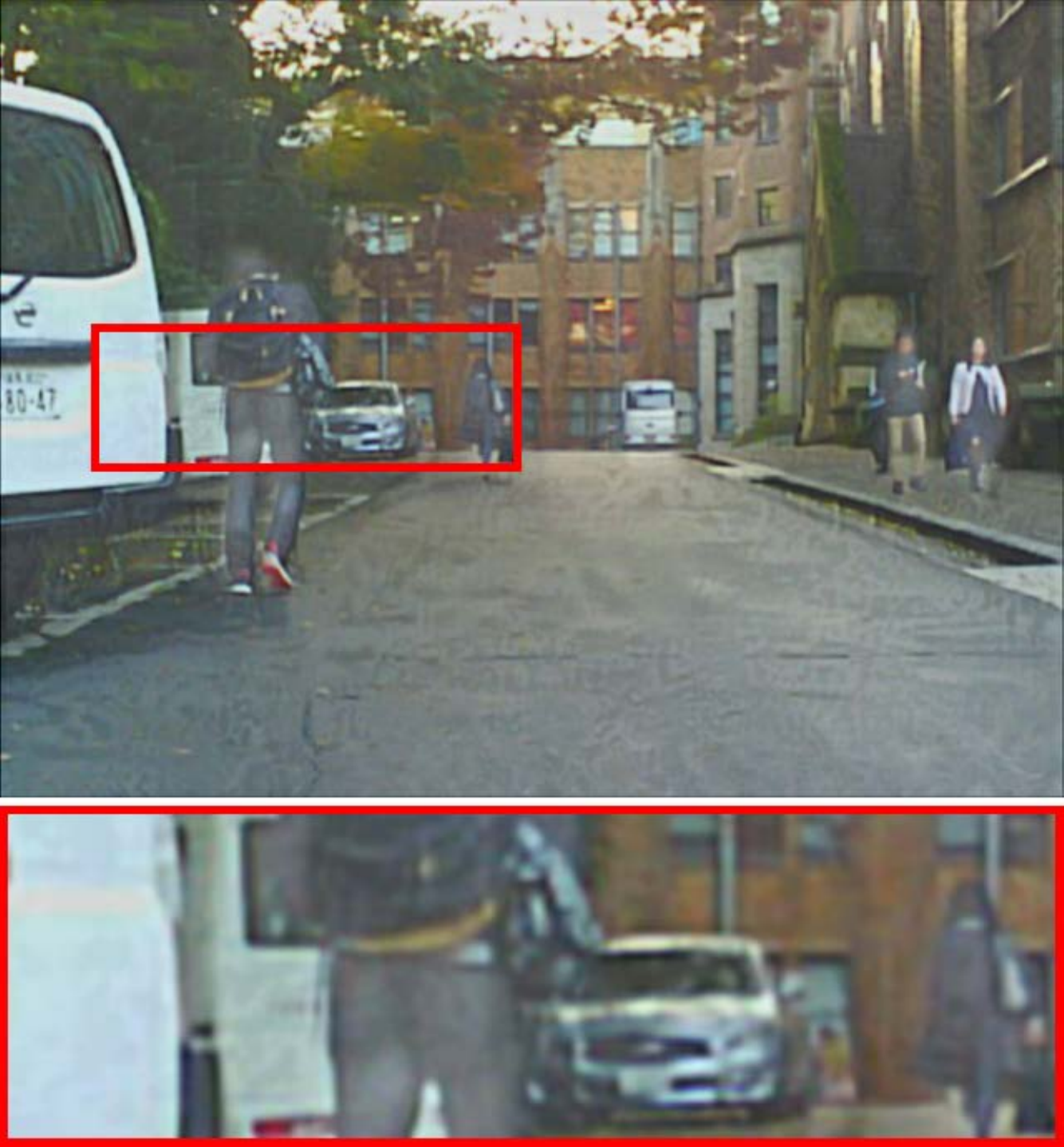}
		&\includegraphics[width=0.135\textwidth]{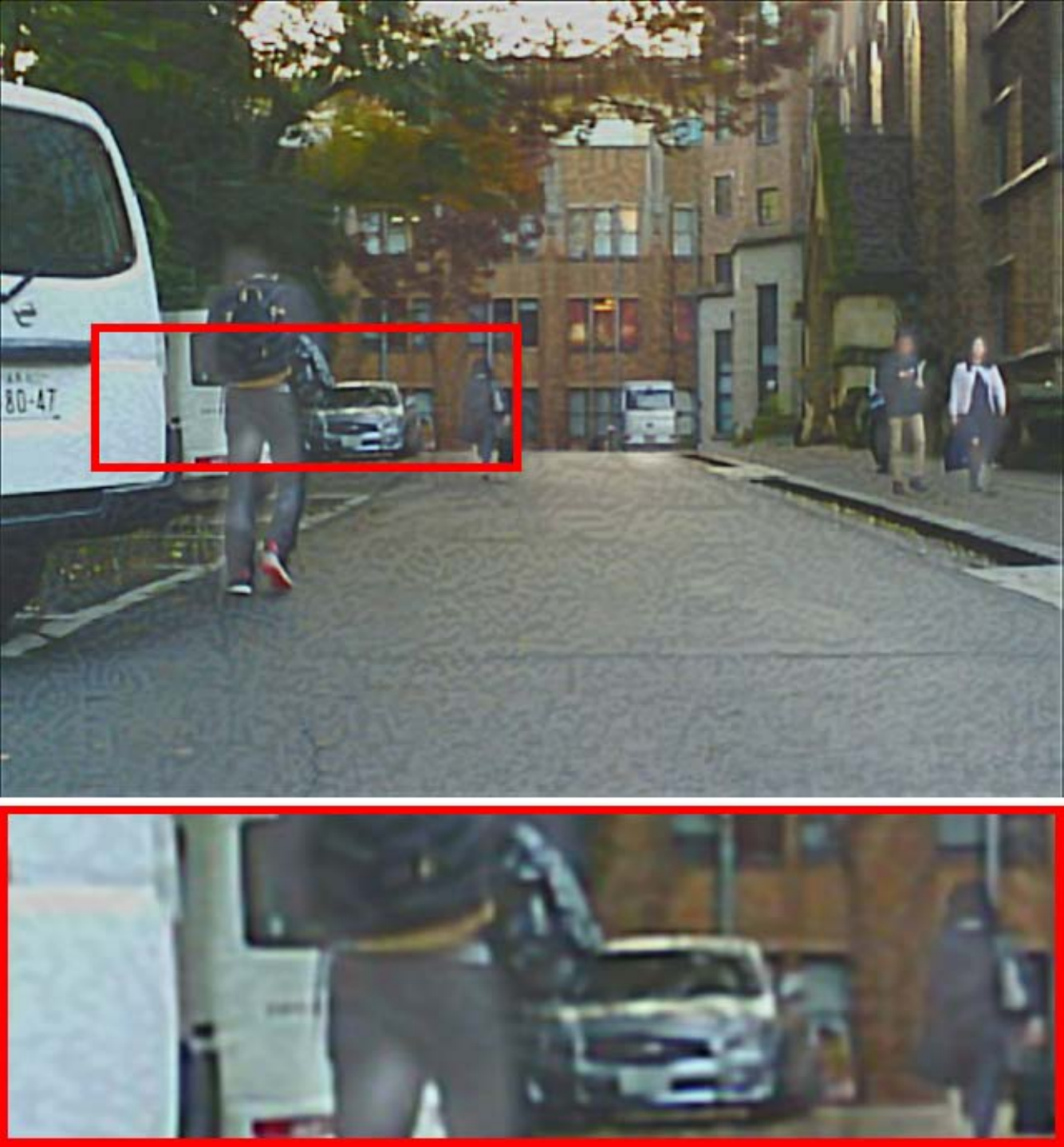}
		&\includegraphics[width=0.135\textwidth]{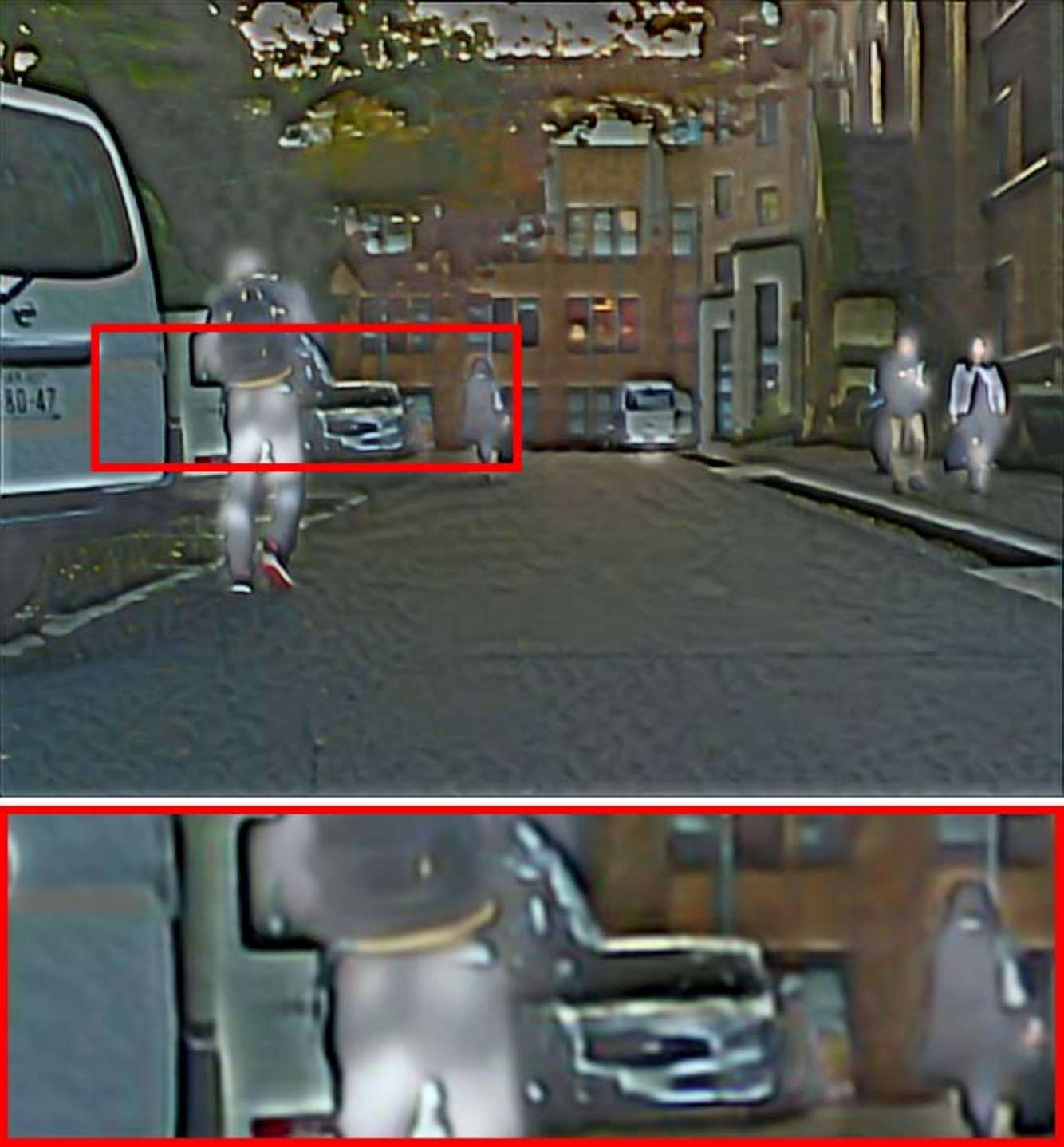}
		&\includegraphics[width=0.135\textwidth]{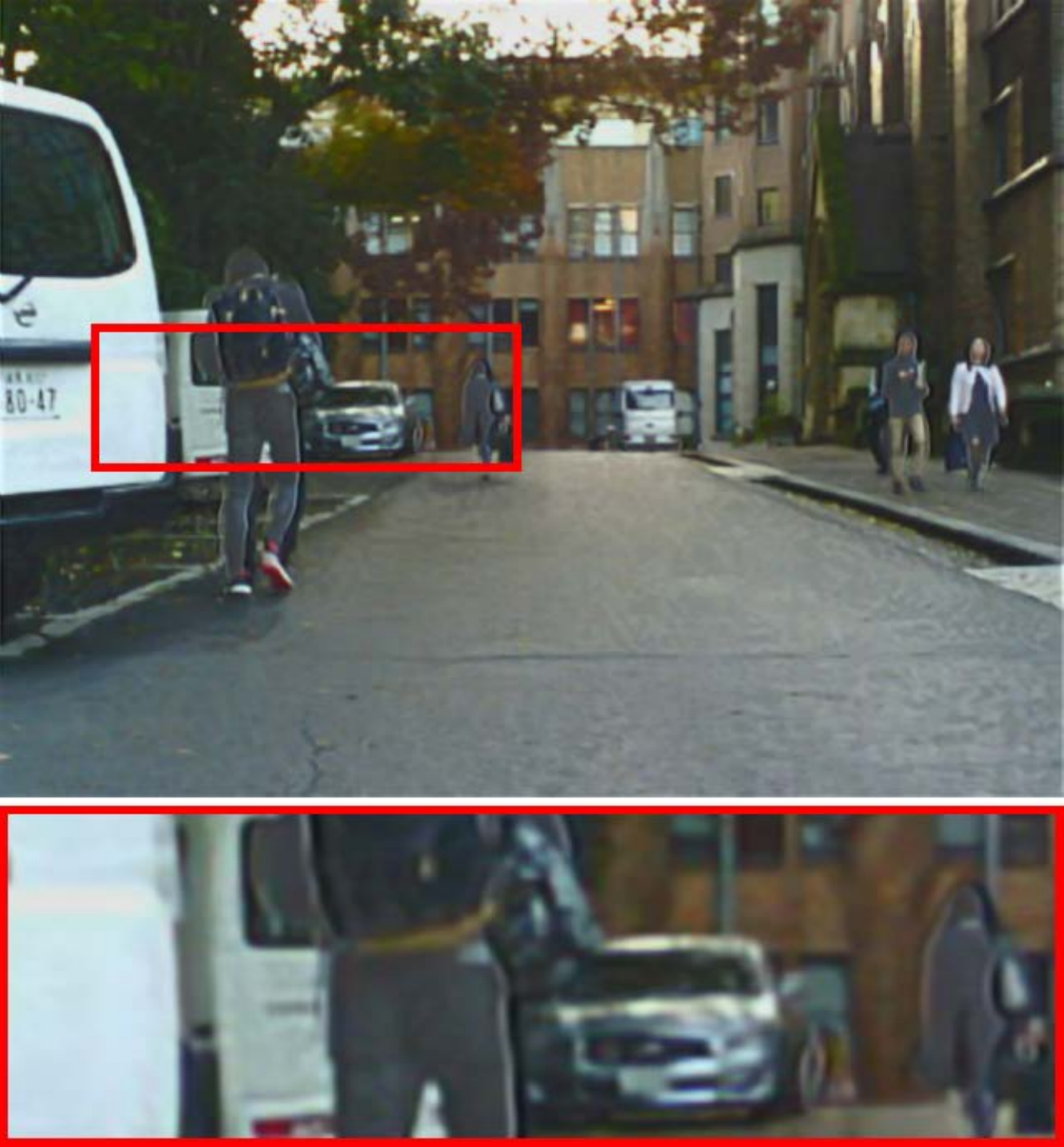}
		&\includegraphics[width=0.135\textwidth]{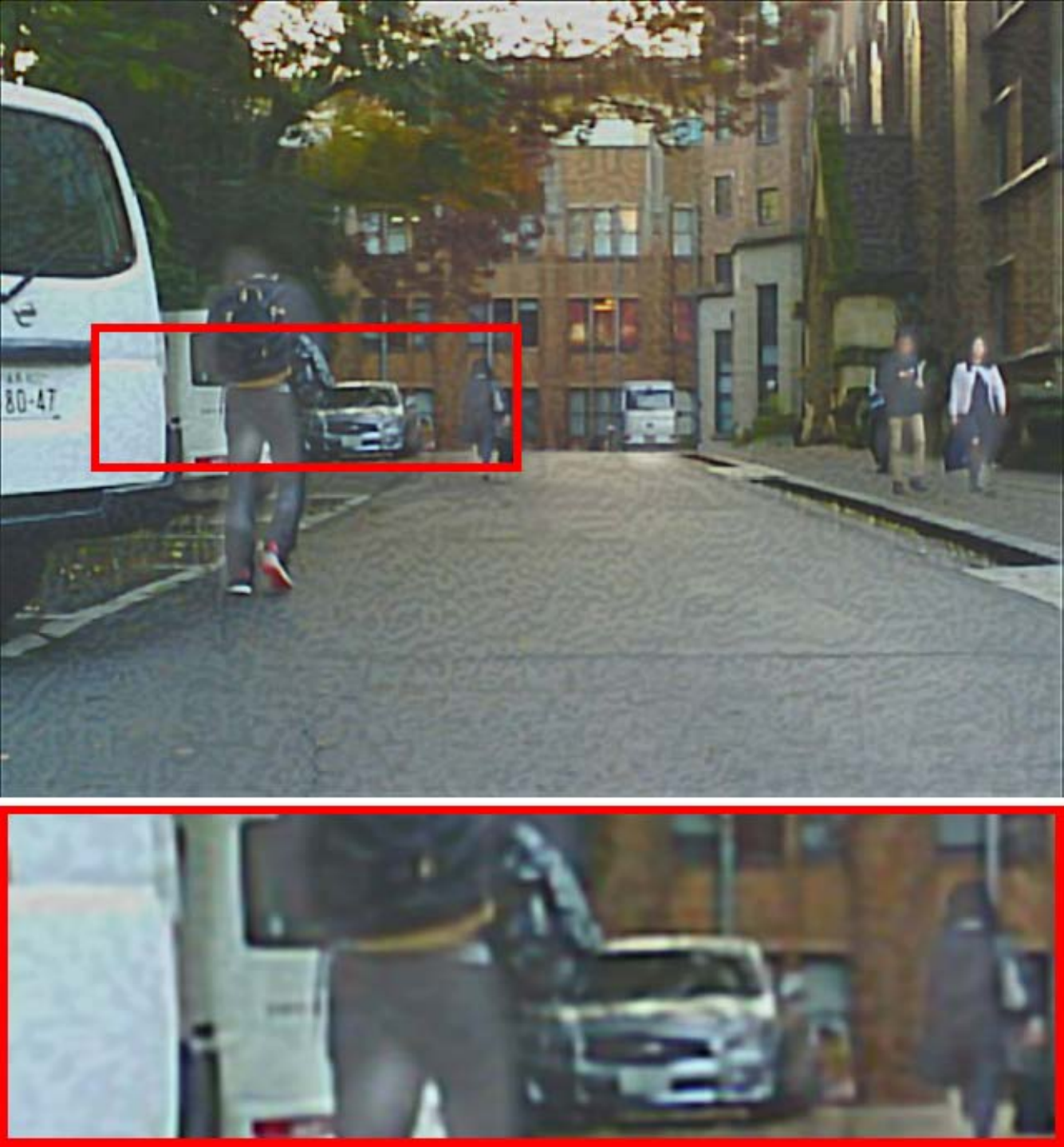}
		&\includegraphics[width=0.135\textwidth]{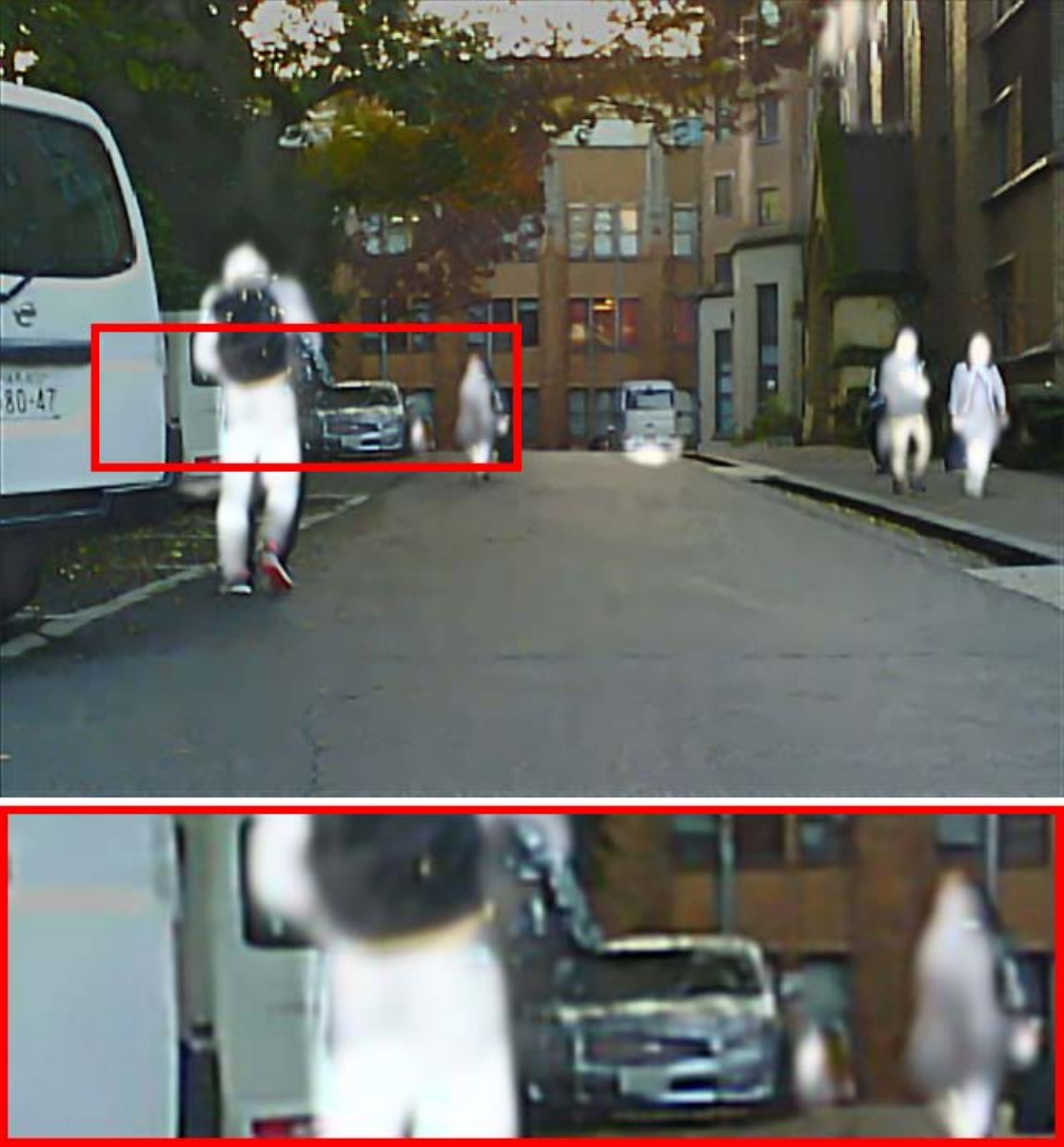}\\

		\footnotesize	Clean Inputs &\footnotesize ReCoNet&\footnotesize UMFusion&\footnotesize DIDFuse&\footnotesize U2Fusion&\footnotesize TarDAL& \footnotesize Ours
		\\
	\end{tabular}
	\vspace{-1em}
	\caption{Visual comparisons to analyse the robustness of proposed architectures with existing advanced networks under the similar adversarial attacks (PGD with $\bm{\bm{\epsilon}} = 8/255$).}
	\label{fig:fusion}
\end{figure*}
\begin{figure}[htb]
	\centering
	\includegraphics[width=0.48\textwidth]{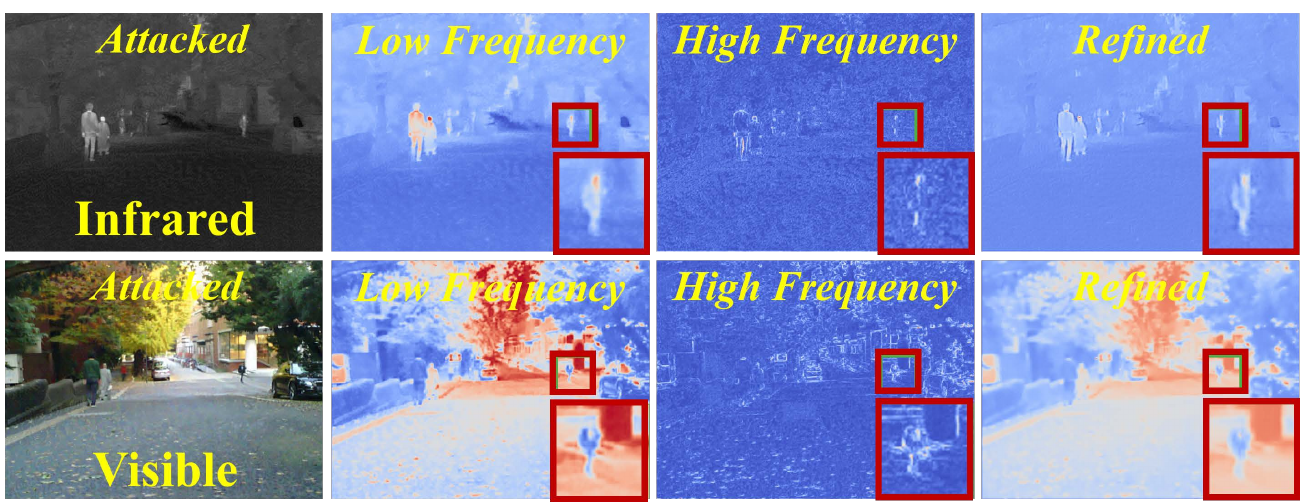}
	
	\caption{{Low-frequency, high-frequency and refined features extracted by the proposed fusion network.}}
	\label{fig:decom}
\end{figure}
\section{Experiments}
\subsection{Implementation Details}

\textit{Loss functions.} We utilize two categories of losses to composite the training and validation loss.
In order to preserve the typical characteristics from source modality, we leverage the visual saliency mechanism~\cite{ma20171123} to measure the typical intensities. Denoted the salient maps as $\mathbf{M}_{x}$ and $\mathbf{M}_{y}$, we utilize  the MSE loss and SSIM loss to define the fusion loss $\mathcal{L}_\mathtt{F}$, which can be formulated as: 
\begin{equation}
\mathcal{L}_\mathtt{F} = \|\mathbf{u}-\mathbf{M}_{x}\mathbf{x}\|_{2}+\|\mathbf{u}-\mathbf{M}_{y}\mathbf{y}\|_{2}+ \text{SSIM}(\mathbf{u};(\mathbf{M}_{x}\mathbf{x}+\mathbf{M}_{y}\mathbf{y})).\label{eq:fusionloss}
\end{equation}
As for the specific semantic perception tasks, we introduce the task-specific losses (denoted as $\mathcal{L}_\mathtt{T}$). Semantic segmentation network is composited by Segformer~\cite{xie2021segformer} with cross-entropy loss. We utilize the FCOS detector~\cite{tian2019fcos} with hybrid losses to conduct the multi-modality detection task. The average combination of $\mathcal{L}_\mathtt{F}$ and $\mathcal{L}_\mathtt{T}$ composites the loss $\mathcal{L}_\mathtt{tr}$ and $\mathcal{L}_\mathtt{val}$.

\textit{Search configurations.} As summarized in Alg.~\ref{alg:framework}, we first pre-trained the whole framework $\mathcal{N}\circ\mathcal{T}$ jointly (\textit{i.e.,} including the candidate fusion super-net and segmentation network), aiming to acquire the well-initialized parameters $\bm{\theta}$ and
$\bm{\omega}$ respectively.  Specifically, leveraging the Adamw optimizer with    $8e^{-5}$, we warm-started the search procedure. Then we adopt the PGD strategy to generate adversarial samples with $\bm{\bm{\epsilon}} = 2/255$ in three iterations. We used one-quarter of the adversarial samples and three-quarters of the normal data to form the training and test sets. By introducing the highly efficient solution~\cite{liu2022revisiting}, we updated the relaxed architecture $\bm{\alpha}$ after  each five steps of parameter learning with SGD optimizer and learning rate $5e^{-3}$. With batch size of 4, we searched the fusion architecture with $10^{5}$ iterations.
The experiments were conducted with the PyTorch framework and on an NVIDIA Tesla V100 GPU. The final architecture is composited by 3-$\mathbf{DB}$, 3-$\mathbf{DC}$, 3-$\mathbf{DB}$, 3-$\mathbf{DB}$, $\mathbf{CA}$ and 7-$\mathbf{RB}$ respectively.


\textit{Training configurations.} According to the transferred attacks~\cite{xie2019improving}, we consider the proposed cascaded scheme as the target model, which can defend the off-line adversarial attacks from the source model that attempts to fool. We leveraged the different intensities of PGD (\textit{i.e.,} $\bm{\bm{\epsilon}} = 1/255$,
$2/255$ and $4/255$)  and randomly select 200 pairs from MFNet dataset to generate the small, moderate the heavy attacks, respectively.  Ten steps of gradient update with learning rate $1e^{-4}$ were performed to the optimization of each attack (lines 3-4 in Alg.~\ref{alg:training}). Then we optimized the initial weights $\bm{\theta}$ (lines 6 in Alg.~\ref{alg:training}) using learning rate $1e^{-3}$. Then we introduced the standard  training exploiting equally divided  normal  and adversarial data with 5 steps of PGD ($\bm{\epsilon}=8/255$). Datasets including (MFNet~\cite{takumi2017multispectral} for segmentation, and M3FD~\cite{TarDAL}  for detection) are utilized.

We compare the comprehensive validation (qualitative and quantitative comparisons) with the advanced image fusion methods, including U2Fusion~\cite{U2Fusion2020}, DIDFuse~\cite{zhao2020didfuse}, ReCoNet~\cite{reconet}, UMFusion~\cite{UMFusion}, TarDAL~\cite{TarDAL}, SeaFusion~\cite{SeaFusion} and  SDNet~\cite{zhang2021sdnet}.
In order to perform a fair comparison, we train these fusion approaches with the same segmentation network under the same loss functions.


\subsection{Infrared-Visible Segmentation Results}

\textit{Qualitative comparison.} We select two PGD attack strategies to illustrate the remarkable performance compared with seven methods in Fig.~\ref{fig:segcomp}. Two significant advantages can be concluded from these instances. Firstly, our scheme can effectively preserve the complete structure of thermal-sensitive targets (\textit{e.g.}, pedestrian) with precise prediction under the attack of PGD. Secondly, our scheme also highlights abundant texture-salient information for the thermal-insensitive categories (\textit{e.g.,} car and bike). Obviously, our method can precisely estimate the structure of cars and bikes on two  images under diverse adversarial perturbations.

\textit{Quantitative comparison.} Table~\ref{tab:segmentation} provides detailed numerical results for five categories of the MFNet dataset under various attack strategies. Our method exhibits superior robustness compared to other comparison methods against diverse attacks. For instance, our scheme improves the mIOU by 15.3\% and 24.6\% under diverse adversarial levels, respectively. Furthermore, our method outperforms other comparative methods on clean data as well, which also can outperform the lastest dual-stream segmentation methods~\cite{lasnet,zhou2022edge}. Specifically, compared to SeaFusion, a segmentation-driven method, our scheme significantly enhances the mIOU by 26.2\%.

\subsection{Extension to  Object Detection}

\textit{Qualitative comparison.} As shown in Fig.~\ref{fig:detec}, we select two challenging scenarios including extreme darkness  and severe degradation to illustrate our remarkable  performance. We also apply imperceptible adversarial perturbations (\textit{i.e.,} PGD with $\bm{\epsilon}=1/255$). Obviously, even if these methods are trained in an adversarial defence way, they cannot accurately detect all targets of the scenes, affected by multiple degradation factors. For instance, as shown in the second row, most methods cannot detect vehicle information correctly due to the interference of dense fog and strong light. In contrast, our method is able to maintain consistent stability of detection in these harsh scenarios.

\textit{Quantitative comparison.} Table.~\ref{tab:detec} reports the detection performance under three scenarios (\textit{e.g.,} clean dataset, small and moderate attacks). Our solutions are significantly  robuster to these adversarial environments (promoting 23.6\% and 23\% mAP) while maintaining extremely high detection accuracy.
We also provide the concrete model size, FLOPs, and inference time of fusion networks. We emphasize that, maintaining extremely high computational efficiency, the proposed fusion network can also provide robust visual features for subsequent perception tasks.
\begin{figure}[htb]
	\centering \begin{tabular}{c@{\extracolsep{0.15em}}c@{\extracolsep{0.15em}}c@{\extracolsep{0.15em}}c}
		\includegraphics[width=0.11\textwidth]{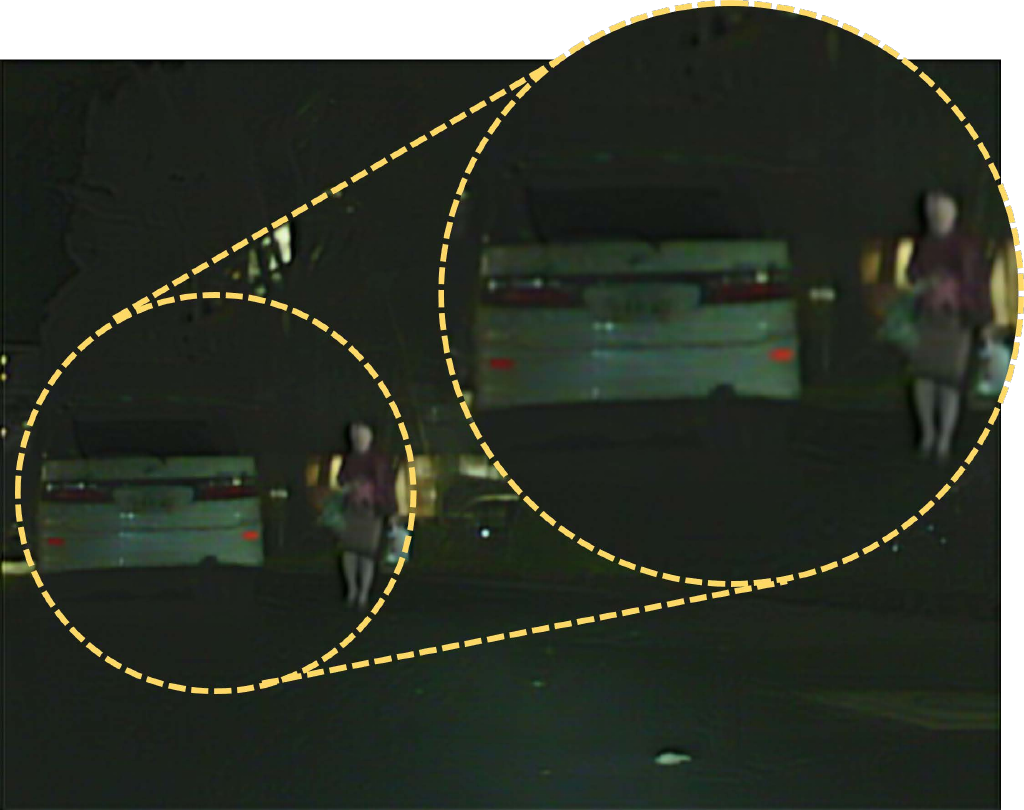}
		& \includegraphics[width=0.11\textwidth]{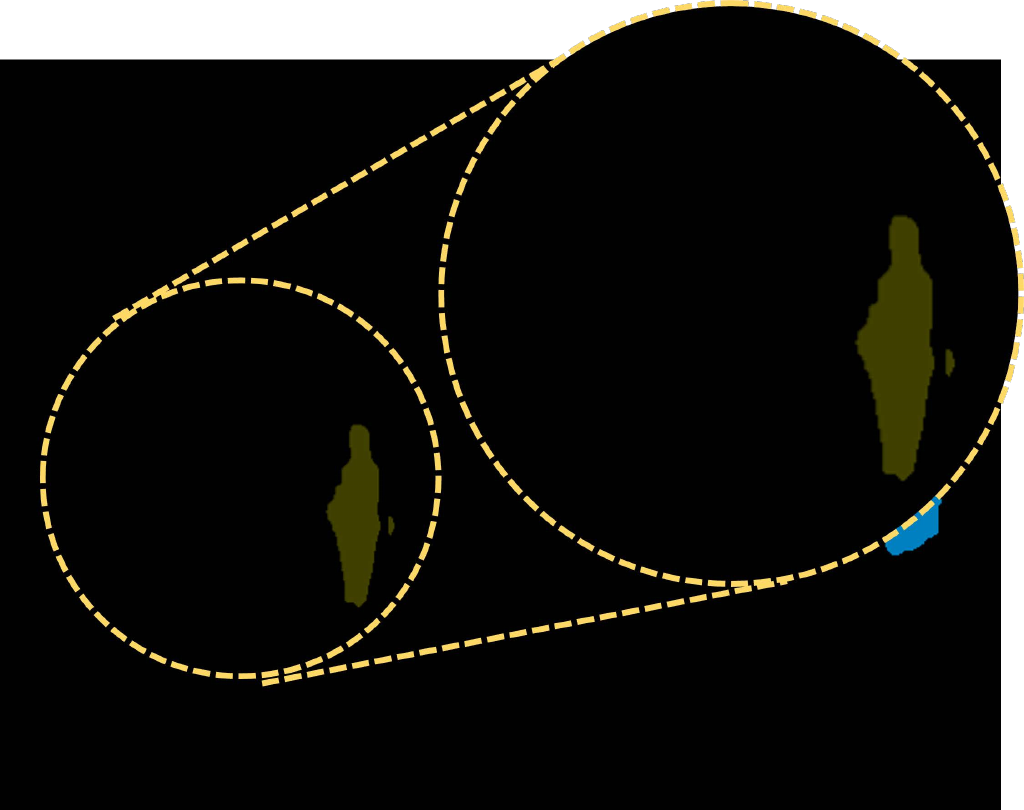}
		&		\includegraphics[width=0.11\textwidth]{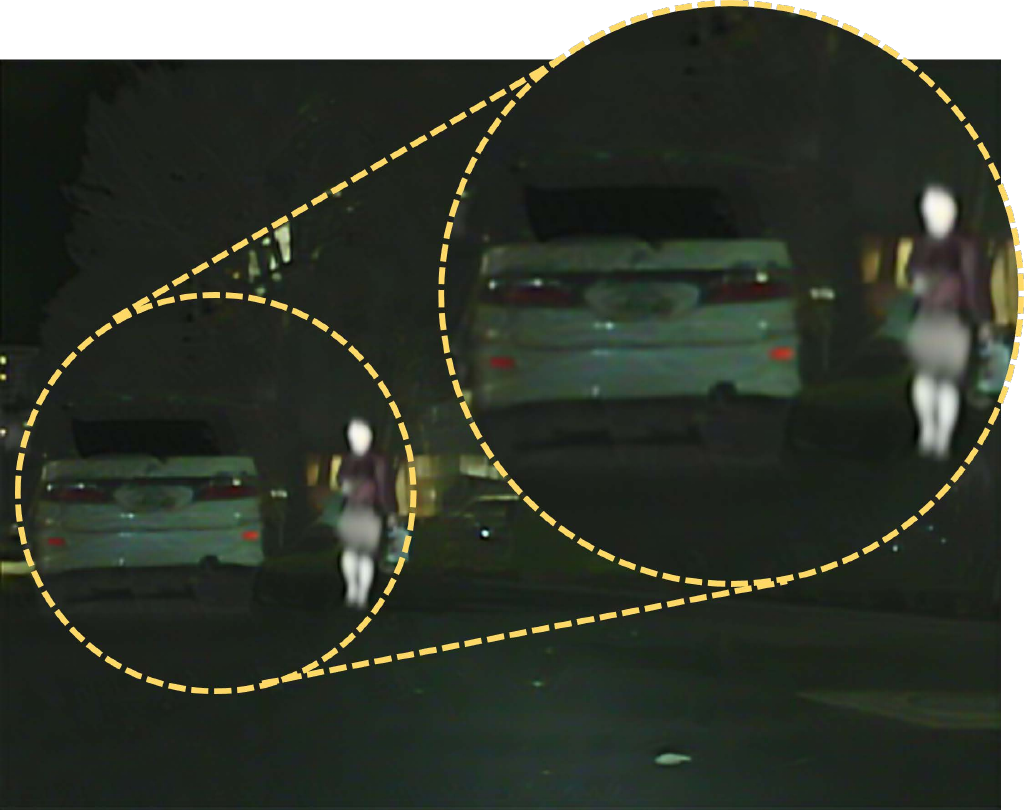}
		
		&		\includegraphics[width=0.11\textwidth]{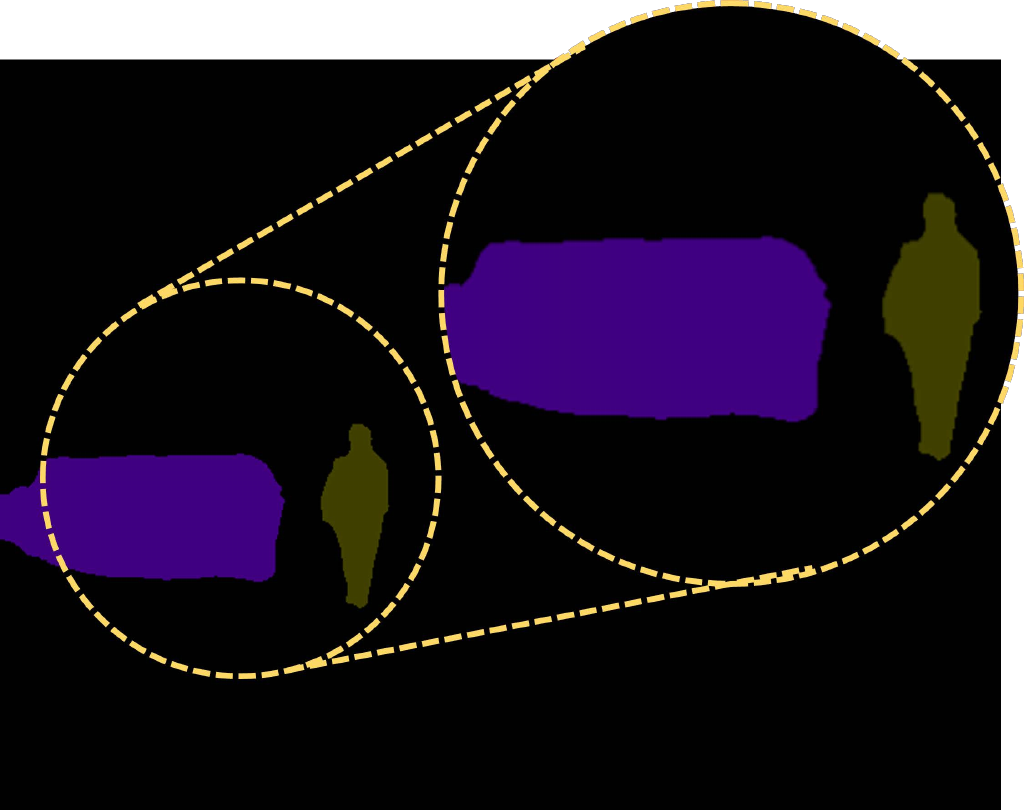}\\
		\includegraphics[width=0.11\textwidth]{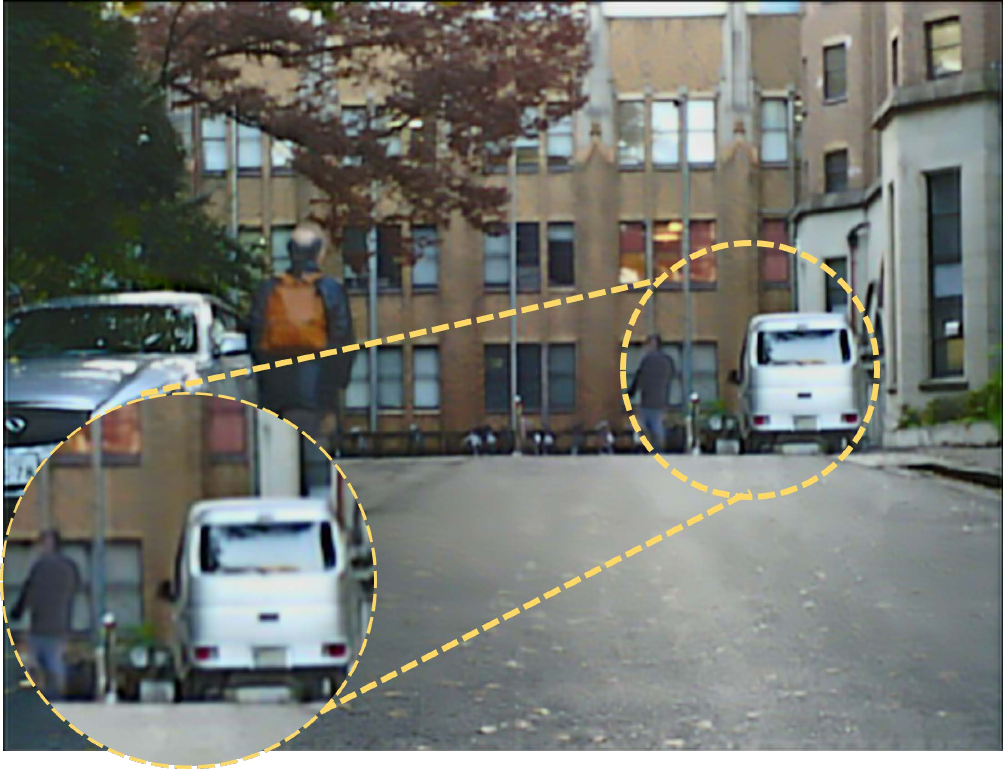}
		& \includegraphics[width=0.11\textwidth]{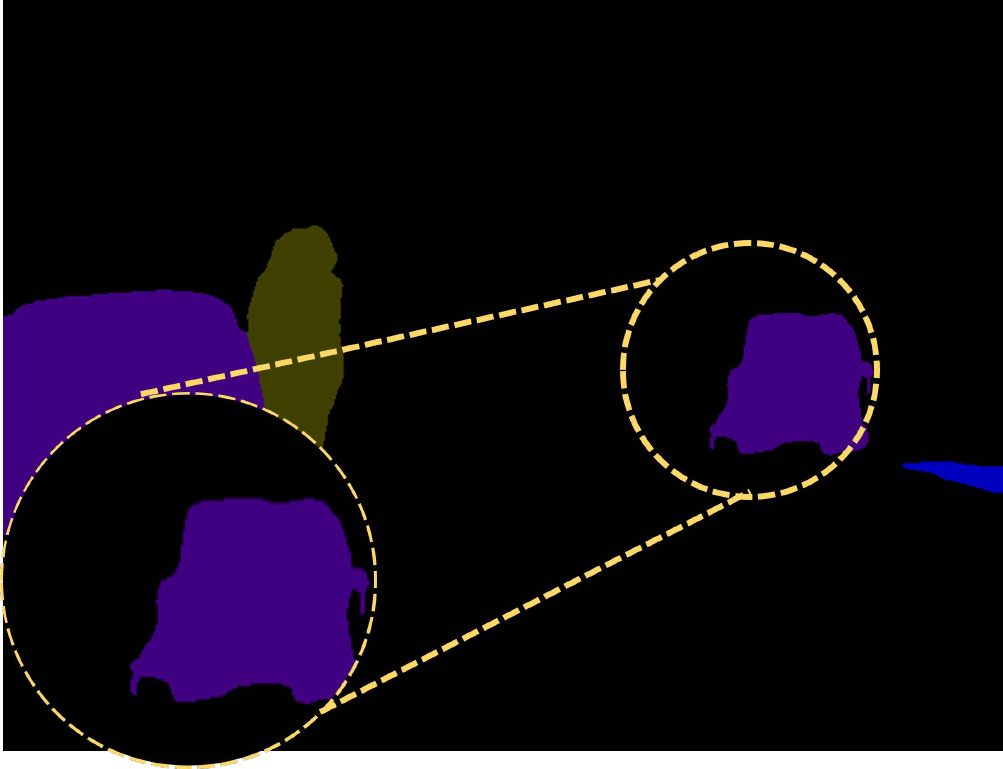}
		&		\includegraphics[width=0.11\textwidth]{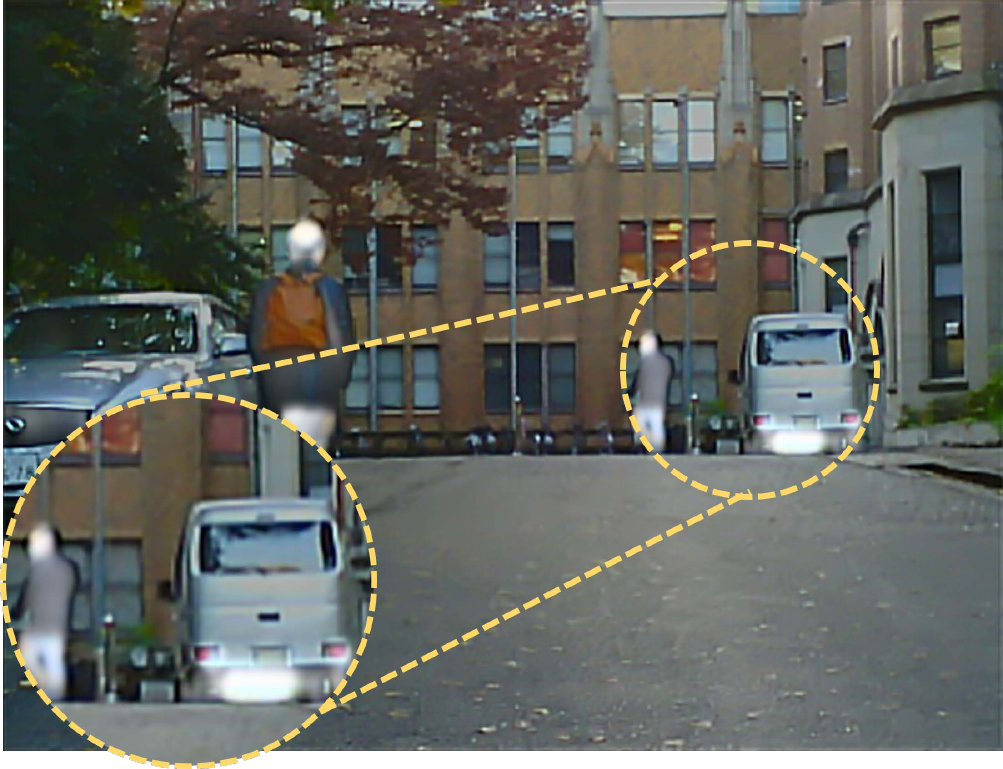}
		
		&		\includegraphics[width=0.11\textwidth]{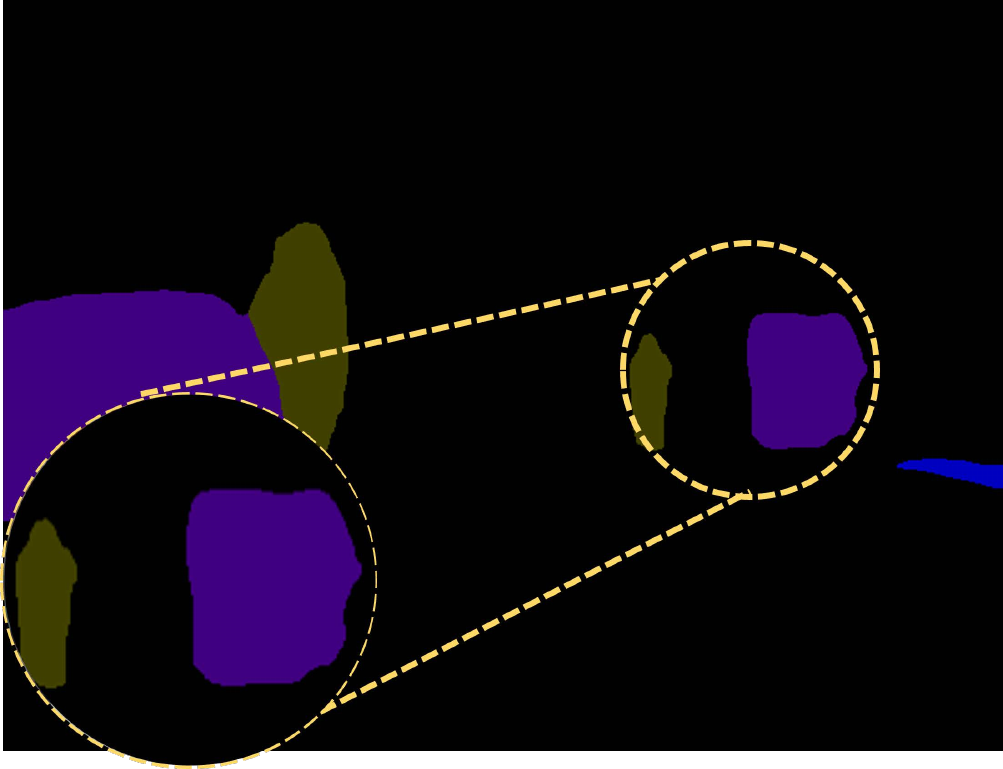}\\
		\multicolumn{2}{c}{\footnotesize(a) SAT} &\multicolumn{2}{c}{\footnotesize(b) AAT} 
	\end{tabular}
	\vspace{-0.4cm}
	\caption{Visual comparisons of different training strategies with fused images and segmentation results (PGD $\bm{\bm{\epsilon}} = 4/255$).}
	\label{fig:training}
\end{figure}
\begin{table*}[htb]
	\centering
	\caption{Validating the effectiveness of training strategies on semantic segmentation task.}~\label{tab:segmentation}
	
	\renewcommand{\arraystretch}{1.0}
	\vspace{-0.4cm}
	\setlength{\tabcolsep}{0.3mm}{
		\begin{tabular}{c|cccccc|cccccc|cccccc}
			\hline
			\multirow{2}{*}{Strategy} & \multicolumn{6}{c|}{Clean Images}                                                                                                                         & \multicolumn{6}{c|}{
				PGD ($\bm{\bm{\epsilon}} = 4/255$) }                                                                                                                                               & \multicolumn{6}{c}{PGD ($\bm{\bm{\epsilon}} = 8/255$)}                                                                                                                      \\ \cline{2-19} 
			& \multicolumn{1}{c|}{\cellcolor{gray!20} Car} & \multicolumn{1}{c|}{\cellcolor{gray!20}Person} & \multicolumn{1}{c|}{\cellcolor{gray!20}Bike} & \multicolumn{1}{c|}{\cellcolor{gray!20}Cone} & \multicolumn{1}{c|}{\cellcolor{gray!20}Bump} & \cellcolor{gray!20} mIOU $\uparrow$ & \multicolumn{1}{c|}{\cellcolor{gray!20} Car} & \multicolumn{1}{c|}{\cellcolor{gray!20} Person} & \multicolumn{1}{c|}{\cellcolor{gray!20} Bike} & \multicolumn{1}{c|}{\cellcolor{gray!20} Cone} & \multicolumn{1}{c|}{\cellcolor{gray!20}Bump} & \multicolumn{1}{c|}{\cellcolor{gray!20} mIOU $\uparrow$ } & \multicolumn{1}{c|}{\cellcolor{gray!20} Car} & \multicolumn{1}{c|}{\cellcolor{gray!20}Person} & \multicolumn{1}{c|}{\cellcolor{gray!20} Bike} & \multicolumn{1}{c|}{\cellcolor{gray!20} Cone} & \multicolumn{1}{c|}{\cellcolor{gray!20} Bump} & \cellcolor{gray!20} mIOU $\uparrow$  \\ \hline
			General	& \multicolumn{1}{c|}{\textbf{0.881}}    & \multicolumn{1}{c|}{\textbf{0.724}}       & \multicolumn{1}{c|}{0.608}     & \multicolumn{1}{c|}{\textbf{0.560}}     & \multicolumn{1}{c|}{\textbf{0.572}}     & \textbf{0.565}      
			& \multicolumn{1}{c|}{0.277}    & \multicolumn{1}{c|}{0.172}       & \multicolumn{1}{c|}{0.174}     & \multicolumn{1}{c|}{0.068}     & \multicolumn{1}{c|}{0.006}     &         0.179                 
			& \multicolumn{1}{c|}{0.256}    & \multicolumn{1}{c|}{0.121}       & \multicolumn{1}{c|}{0.124}     & \multicolumn{1}{c|}{0.038}     & \multicolumn{1}{c|}{0.000}     &  0.157    \\ \hline
			SAT	& \multicolumn{1}{c|}{0.835}    & \multicolumn{1}{c|}{0.673}       & \multicolumn{1}{c|}{0.619}     & \multicolumn{1}{c|}{0.424}     & \multicolumn{1}{c|}{0.406}     &   0.507   
			& \multicolumn{1}{c|}{\textbf{0.665}}    & \multicolumn{1}{c|}{0.441}       & \multicolumn{1}{c|}{0.338}     & \multicolumn{1}{c|}{0.256}     & \multicolumn{1}{c|}{0.006}     &         0.328                 
			& \multicolumn{1}{c|}{0.590}    & \multicolumn{1}{c|}{0.294}       & \multicolumn{1}{c|}{0.227}     & \multicolumn{1}{c|}{0.184}     & \multicolumn{1}{c|}{0.000}     &  0.267    \\ \hline
			AAT	& \multicolumn{1}{c|}{0.866}    & \multicolumn{1}{c|}{0.712}       & \multicolumn{1}{c|}{\textbf{0.634}}     & \multicolumn{1}{c|}{0.504}     & \multicolumn{1}{c|}{0.312}     & 0.518  
			& \multicolumn{1}{c|}{0.643}    & \multicolumn{1}{c|}{\textbf{0.563}}       & \multicolumn{1}{c|}{\textbf{0.437}}     & \multicolumn{1}{c|}{\textbf{0.278}}     & \multicolumn{1}{c|}{\textbf{0.041}}     &      \textbf{0.361}            
			& \multicolumn{1}{c|}{\textbf{0.603}}    & \multicolumn{1}{c|}{\textbf{0.466}}       & \multicolumn{1}{c|}{\textbf{0.329}}     & \multicolumn{1}{c|}{\textbf{0.253}}     & \multicolumn{1}{c|}{0.000}     &  \textbf{0.314}    \\ \hline
			
		\end{tabular}
		
	}
\end{table*}
\subsection{Visual Comparison of Image Fusion}
We also analyze the robustness of fusion with advanced competitors from the visual perspective, which is shown in Fig.~\ref{fig:fusion}. Note that we do not compare the fusion performance based on the general statistic metrics, which only measure the source information maintenance, easy to be interfered with by adversarial artifacts. Compared with these existing methods, which are trained with standard adversarial defence, our scheme has two significant advantages. Firstly, the comprehensive characteristics can be effectively preserved. The discriminative targets (\textit{e.g.,} the structure of car and pedestrian) from infrared modality are highlighted from the strong light interference. Furthermore, our approach can significantly remove the adversarial artifacts. 
\section{Ablation Study}

\textit{Effects of search space.} We first evaluated the effectiveness of searched architecture with single-operation composited networks under standard adversarial training and reported the experimental results in  Table.~\ref{tab:operations}. As above mentioned, 3-$\mathbf{DC}$ and $\mathbf{CA}$ realize consistent robustness. However, there still exists an obvious gap compared with the finally searched networks. 

\begin{table}[]
	\centering
	
	\caption{Effectiveness of operations on MFNet benchmark.}~\label{tab:operations}
	
	\renewcommand{\arraystretch}{1.1}
	\setlength{\tabcolsep}{1.3mm}{
		\vspace{-0.4cm}
		\begin{tabular}{c|ccc|ccc}
			\hline
			\multirow{2}{*}{Operations} &\multicolumn{3}{c|}{PGD ($\bm{\bm{\epsilon}} = 4/255$)}                                                             & \multicolumn{3}{c}{PGD ($\bm{\bm{\epsilon}} = 8/255$)}                                             \\ \cline{2-7} 
			& \multicolumn{1}{c|}{\cellcolor{gray!20}Car} & \multicolumn{1}{c|}{\cellcolor{gray!20}Person} & \multicolumn{1}{c|}{\cellcolor{gray!20}mIOU$\uparrow$} & \multicolumn{1}{c|}{\cellcolor{gray!20}Car} & \multicolumn{1}{c|}{\cellcolor{gray!20}Person} &\cellcolor{gray!20} mIOU$\uparrow$ \\ \hline
			3DC	& \multicolumn{1}{c|}{0.589}    & \multicolumn{1}{c|}{0.405}       &      0.313                     & \multicolumn{1}{c|}{0.438}    & \multicolumn{1}{c|}{0.253}       &   0.249   \\ \hline
			3DB	& \multicolumn{1}{c|}{0.496}    & \multicolumn{1}{c|}{0.441}       &    0.277                       & \multicolumn{1}{c|}{0.336}    & \multicolumn{1}{c|}{0.310}       &  0.211    \\ \hline
			7RB	& \multicolumn{1}{c|}{0.593}    & \multicolumn{1}{c|}{0.347}       &     0.296                      & \multicolumn{1}{c|}{0.386}    & \multicolumn{1}{c|}{0.185}       &  0.218    \\ \hline
			CA	& \multicolumn{1}{c|}{0.515}    & \multicolumn{1}{c|}{\textbf{0.497}}       & 0.302                          & \multicolumn{1}{c|}{0.382}    & \multicolumn{1}{c|}{\textbf{0.363}}       & 0.239     \\ \hline
			Ours	& \multicolumn{1}{c|}{\textbf{0.665} }    & \multicolumn{1}{c|}{{0.441} }       & \textbf{0.328}                         & \multicolumn{1}{c|}{\textbf{0.590}}    & \multicolumn{1}{c|}{{0.294}}       &  \textbf{0.267}  \\ \hline
\end{tabular}}\end{table}

\begin{table}[htb]
	\centering
	\caption{Evaluating of search strategy on segmentation task.}~\label{tab:searchstrategy}
	
	\renewcommand{\arraystretch}{1.1}
	\vspace{-0.4cm}
	\setlength{\tabcolsep}{1.0mm}{
		\begin{tabular}{c|ccc|ccc}
			\hline
			\multirow{2}{*}{Strategy} & \multicolumn{3}{c|}{PGD ($\bm{\bm{\epsilon}} = 4/255$)}                                                             & \multicolumn{3}{c}{PGD ($\bm{\bm{\epsilon}} = 8/255$)}                                  \\ \cline{2-7} 
			& \multicolumn{1}{c|}{\cellcolor{gray!20}Car} & \multicolumn{1}{c|}{\cellcolor{gray!20}Person} & \multicolumn{1}{c|}{\cellcolor{gray!20}mIOU$\uparrow$} & \multicolumn{1}{c|}{\cellcolor{gray!20}Car} & \multicolumn{1}{c|}{\cellcolor{gray!20}Person} & \cellcolor{gray!20}mIOU $\uparrow$\\ \hline
			Original~\cite{liu2018darts}	& \multicolumn{1}{c|}{0.557}    & \multicolumn{1}{c|}{\textbf{0.464}}       &     0.288                      & \multicolumn{1}{c|}{0.347}    & \multicolumn{1}{c|}{0.275}       &   0.208   \\ \hline
			HDS	& \multicolumn{1}{c|}{\textbf{0.665} }    & \multicolumn{1}{c|}{{0.441} }       & \textbf{0.328}                         & \multicolumn{1}{c|}{\textbf{0.590}}    & \multicolumn{1}{c|}{\textbf{0.294}}       &  \textbf{0.267}     \\ \hline
		\end{tabular}
	}
\end{table}

\begin{figure}[htb]
	\centering
	\includegraphics[width=0.48\textwidth]{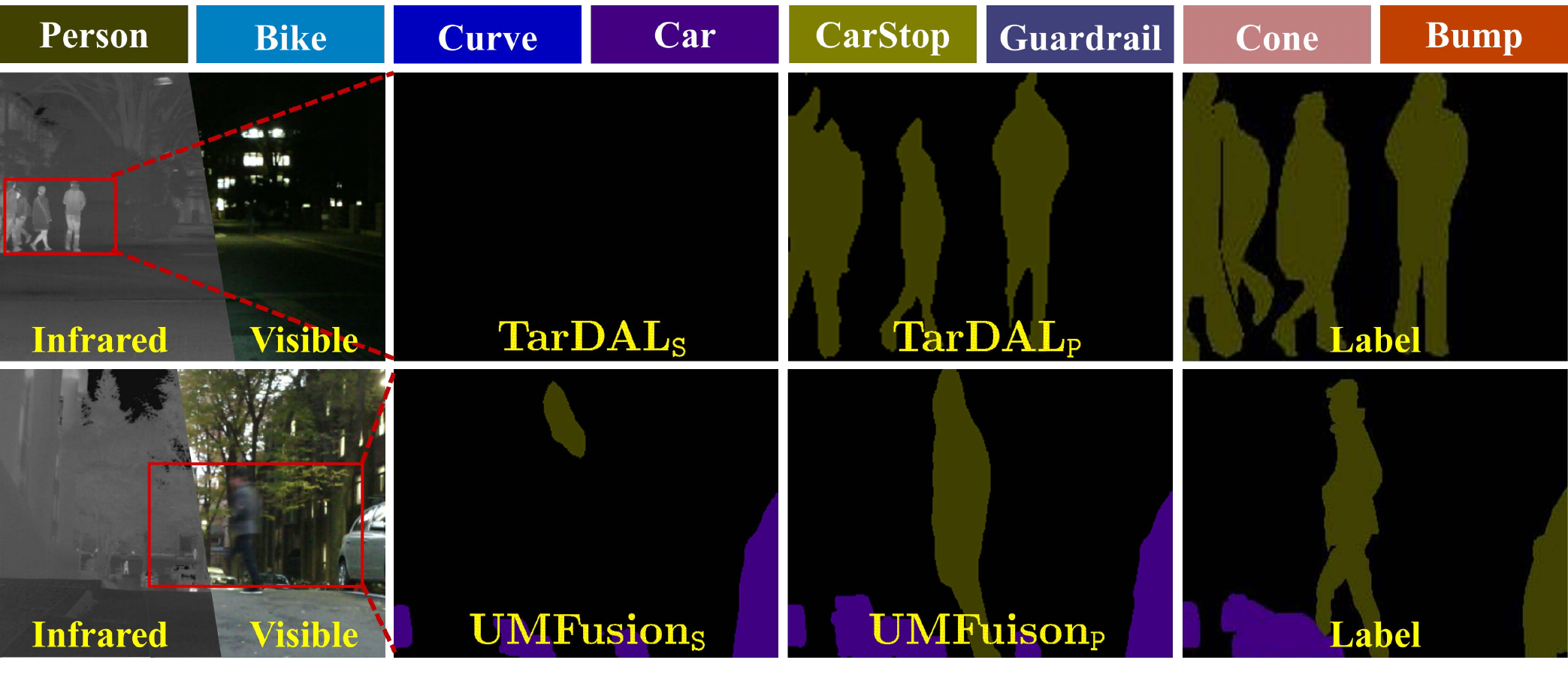}
	\vspace{-0.4cm}
	\caption{{Validating the effectiveness of  proposed training with diverse fusion networks under the adversarial attacks.}}
	\label{fig:segmethod}
\end{figure}

\begin{table}[htb]
	\centering
	\caption{Evaluating the generalization ability of proposed training strategy for advanced fusion architectures.}~\label{tab:generalization}
	\vspace{-0.4cm}
	\renewcommand{\arraystretch}{1.1}
	\setlength{\tabcolsep}{1.4mm}{
		\begin{tabular}{c|ccc|ccc}
			\hline
			\multirow{2}{*}{Strategy}  & \multicolumn{3}{c|}{PGD ($\bm{\bm{\epsilon}} = 4/255$)}                                                           & \multicolumn{3}{c}{PGD ($\bm{\bm{\epsilon}} = 8/255$)}                              \\ \cline{2-7} 
			& \multicolumn{1}{c|}{\cellcolor{gray!20}Car} & \multicolumn{1}{c|}{\cellcolor{gray!20}Person} & \multicolumn{1}{c|}{\cellcolor{gray!20}mIOU$\uparrow$} & \multicolumn{1}{c|}{\cellcolor{gray!20}Car} & \multicolumn{1}{c|}{\cellcolor{gray!20}Person} &\cellcolor{gray!20} mIOU$\uparrow$ \\ \hline
			TarDAL$_\mathtt{S}$	& \multicolumn{1}{c|}{0.538}    & \multicolumn{1}{c|}{0.249}       &  0.264                &  \multicolumn{1}{c|}{0.431}    & \multicolumn{1}{c|}{0.164}       &        0.213          \\ \hline
			TarDAL$_\mathtt{P}$	& \multicolumn{1}{c|}{ \textbf{0.598}}    & \multicolumn{1}{c|}{ \textbf{0.484}}       &  \textbf{0.312}                    & \multicolumn{1}{c|}{\textbf{0.492}  }   & \multicolumn{1}{c|}{\textbf{0.355}  }       & \textbf{0.257}    \\ \hline
			UMFusion$_\mathtt{S}$	& \multicolumn{1}{c|}{\textbf{0.608}}    & \multicolumn{1}{c|}{0.404}       &       0.311                    & \multicolumn{1}{c|}{0.453}    & \multicolumn{1}{c|}{0.199}       &0.233      \\ \hline
			UMFusion$_\mathtt{P}$	& \multicolumn{1}{c|}{0.602}    & \multicolumn{1}{c|}{\textbf{0.525}}       &       {\textbf{0.316}}                    & \multicolumn{1}{c|}{{\textbf{0.454}} }    & \multicolumn{1}{c|}{{\textbf{0.339}}}        & {\textbf{0.245}}     \\ \hline
		\end{tabular}
	}
\end{table}

\begin{figure}[htb]
	\centering
	\includegraphics[width=0.48\textwidth]{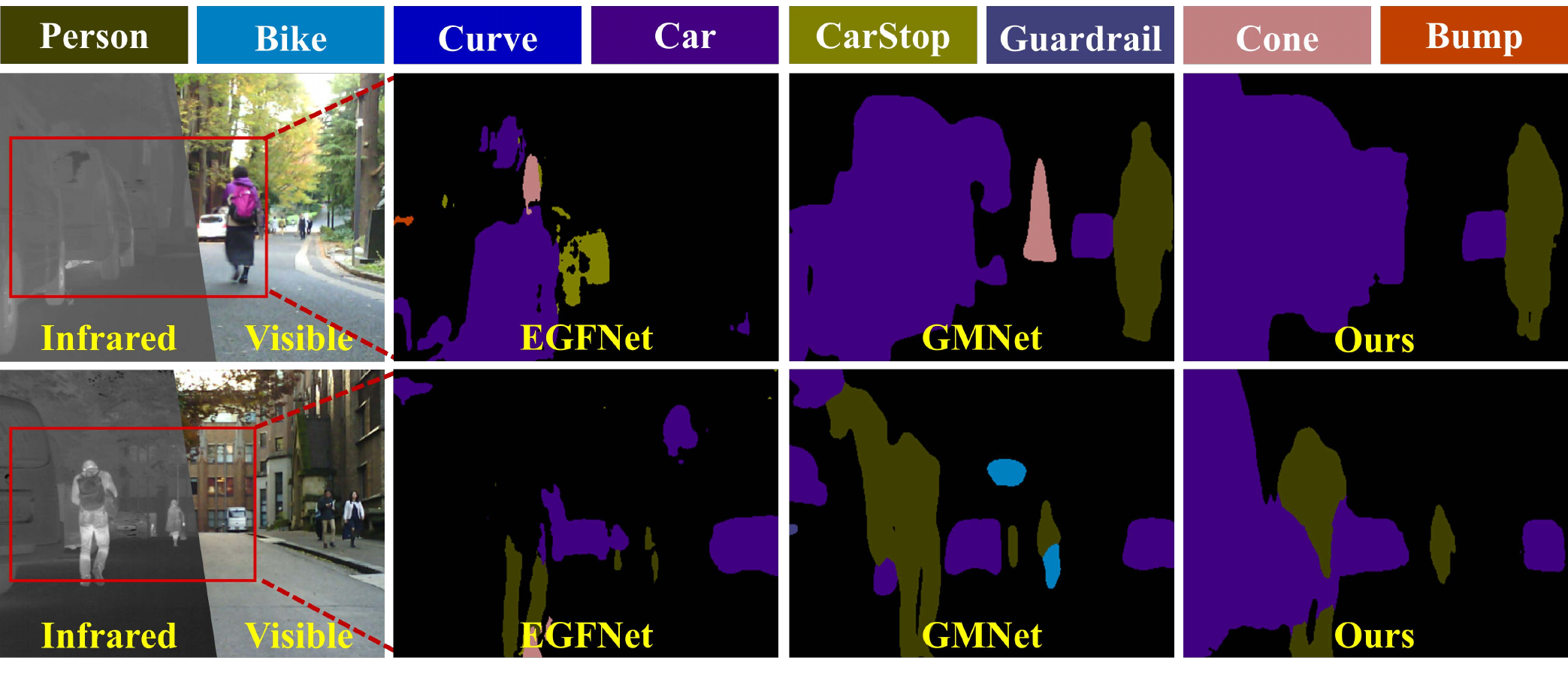}
	
	\caption{{Comparison with  dual-stream segmentation methods under diverse PGD perturbations ($\bm{\bm{\epsilon}} = 4/255$ and $8/255$).}}
	\label{fig:segdual}
\end{figure}
\begin{table}[htb]
	\centering
	\caption{Evaluating the effectiveness of the image fusion module compared with the dual-stream segmentation methods.}~\label{tab:segnetwork}
	
	\renewcommand{\arraystretch}{1.1}
	\vspace{-0.4cm}
	\setlength{\tabcolsep}{1.0mm}{
		\begin{tabular}{c|ccc|ccc}
			\hline
			\multirow{2}{*}{Strategy}  & \multicolumn{3}{c|}{PGD ($\bm{\bm{\epsilon}} = 4/255$)}                                                           & \multicolumn{3}{c}{PGD ($\bm{\bm{\epsilon}} = 8/255$)}                        \\ \cline{2-7} 
			& \multicolumn{1}{c|}{\cellcolor{gray!20}Car} & \multicolumn{1}{c|}{\cellcolor{gray!20}Person} & \multicolumn{1}{c|}{\cellcolor{gray!20}mIOU$\uparrow$} & \multicolumn{1}{c|}{\cellcolor{gray!20}Car} & \multicolumn{1}{c|}{\cellcolor{gray!20}Person} & \cellcolor{gray!20} mIOU $\uparrow$\\ \hline
			LASNet$_\mathtt{S}$~\cite{lasnet}	& \multicolumn{1}{c|}{0.094}    & \multicolumn{1}{c|}{0.234}       &    0.146                       & \multicolumn{1}{c|}{0.049}    & \multicolumn{1}{c|}{0.129}       &0.117      \\ \hline
			EGFNet$_\mathtt{S}$~\cite{zhou2022edge}	& \multicolumn{1}{c|}{0.078}    & \multicolumn{1}{c|}{0.108}       &             0.118              & \multicolumn{1}{c|}{0.051}    & \multicolumn{1}{c|}{0.058}       &   0.108   \\ \hline
			GMNet$_\mathtt{S}$~\cite{zhou2021gmnet}	& \multicolumn{1}{c|}{0.242}    & \multicolumn{1}{c|}{0.364}       &   0.186                        & \multicolumn{1}{c|}{0.168}    & \multicolumn{1}{c|}{0.270}       &  0.154    \\ \hline
			Ours	& \multicolumn{1}{c|}{\textbf{0.643}}    & \multicolumn{1}{c|}{\textbf{0.563}}       &      \textbf{0.361}                    & \multicolumn{1}{c|}{\textbf{0.603}}    & \multicolumn{1}{c|}{\textbf{0.466}}       &  \textbf{0.314}    \\ \hline
		\end{tabular}
	}
\end{table}

\textit{Evaluating of decomposition architecture.}  To demonstrate the critical role of decomposition architecture, we visualize the decomposed features in Fig.~\ref{fig:decom}. By exploiting the residual feature Eq~\eqref{eq:resfeature}, the adversarial artifacts can be precisely obtained by the high-frequency components. Then parallel stems can distill and fine the vital features, shown in the last column of Fig.~\ref{fig:decom}.


\textit{Effects of searching strategy.} We also analyze the effectiveness of  search. We compare our strategy with the original differentiable search, which directly utilizes the same loss formulation and one-step approximation. It can be clearly seen that our search scheme improves the network structure and promotes the segmentation performance by 13.8\% under $\bm{\epsilon}=4/255$ perpetuations in Table.~\ref{tab:searchstrategy}.

\textit{Evaluating of training strategy.} 
The proposed  strategy is to offer an attack-tolerant fusion for following perception tasks. We validate the presented training with normal learning and Standard Adversarial Training (SAT) in Table.~\ref{tab:segmentation}. Compared with normal training, the proposed strategy maintain the remarkable numerical results, especially the classification for person and car. Compared with SAT, our schemes realize  better estimation, which can drastically improve 10.1\% and 17.6\% in term of mIOU under perturbations of $\bm{\bm{\epsilon}} = 4/255$ and $\bm{\bm{\epsilon}} = 8/255$ respectively. We also plotted the visual comparisons in Fig.~\ref{fig:training}. Clearly, though artifacts can be mostly removed from both strategies, the proposed scheme can effectively highlight the thermal-sensitive targets (\textit{e.g.,} pedestrians) for the pixel-wise segmentation.

Furthermore, our strategy is network-agnostic, which can be generalized for diverse fusion architectures. Table.~\ref{tab:generalization} reports the improvements of TarDAL and UMFusion under our strategy. We utilize the subscript ``$\mathtt{S}$'' to denote the SAT and ``$\mathtt{P}$'' to represent the adaptive training strategy. Especially for TarDAL, the model can be significantly promoted by 18.2\% and 20.7\% respectively.
An intuitive visual comparison is shown in Fig.~\ref{fig:segmethod}, which contain two challenging instances (\textit{i.e.,} extreme darkness and motion blurs) with severe artifacts. 

\textit{Effectiveness of image fusion.}
We argue that pixel-wise image fusion can effectively aggregate the typical modality characteristics and protect the informative features from adversarial attacks.  In this part, we conducted the standard adversarial training with these schemes and reported the concrete numerical results in Table.~\ref{tab:segnetwork}. Obviously, feature-level fusion, either using element-wise
summation (EGFNet), channel attention  (LASNet) or dense connections (GMNet) are fragile to adversarial attacks. Furthermore,  we also depict the visual differences in Fig.~\ref{fig:segdual}. Under heavy attacks, 
EGFNet fails to predict the scene categories. 
Our method realizes consistent performance.

\section{CONCLUSION}
Within this paper, a perception-aware image fusion framework was proposed to improve robustness for multi-modality semantic segmentation tasks. We first systematically discovered the robust operations and rules of image fusion for segmentation, to eliminate the attack-sensitive
components. Based on the decomposition principle, we designed the harmonized architecture search to automatically explore the perception-aware fusion network. Then we devised the adaptative training strategy to strengthen the parameters robustness of fusion based on diverse transfer-based adversarial attacks. Experiments demonstrate that the proposed scheme realizes outstanding performance consistently on semantic segmentation under diverse  adversarial attacks. Since our framework is generic enough, we can further address other  perception tasks in the future.
\section*{ACKNOWLEDGEMENTS}
This work is partially supported by the National Key R\&D Program of China (No. 2022YFA1004101), the National Natural Science Foundation of China (No. U22B2052).

\newpage
\balance
\bibliographystyle{ACM-Reference-Format}
\bibliography{egbib}

\end{document}